\documentclass[lettersize,journal,9pt]{IEEEtran}

\usepackage[caption=false,font=normalsize,labelfont=sf,textfont=sf]{subfig}
\usepackage{amsmath,amsfonts,siunitx}
\usepackage{algorithmic}
\usepackage{algorithm}
\usepackage{array}
\usepackage{textcomp}
\usepackage{stfloats}
\usepackage{url}
\usepackage{verbatim}
\usepackage{graphicx}
\usepackage{cite}
\usepackage[nolist]{acronym}
\usepackage{amssymb}
\usepackage{orcidlink}
\usepackage{blindtext}
\usepackage{mathtools}
\usepackage{booktabs}
\usepackage[super]{nth}
\usepackage{subfig}
\usepackage{hyperref}
\hypersetup{
    hidelinks
}
\DeclarePairedDelimiter{\abs}{\lvert}{\rvert}

\makeatletter
\renewcommand{\thefigure}{\num{\arabic{figure}}}
\makeatother

\usepackage{comment}
\usepackage{pgfplots}
\pgfplotsset{compat=1.18}

\begin{document}

\begin{acronym}[]	
\acro{vpa}[VPA]{velocity profile analysis}
\acro{ransac}[RANSAC]{random sample consensus algorithm}
\acro{ros2}[ROS2]{Robot Operating System 2}
\acro{fmcw}[FMCW]{Frequency Modulated Continuous Wave}
\acro{if}[IF]{intermediate frequency}
\acro{awgn}[AWGN]{additive white Gaussian noise}
\acro{adas}[ADAS]{Advanced Driver Assistance Systems}
\acro{ad}[AD]{Autonomous Driving}
\acro{dbscan}[DB-SCAN]{Density-Based Spatial Clustering of Applications with Noise}
\acro{FoV}[FoV]{Field-of-View}
\acro{IPS}[IPS]{Indoor Position System}
\acro{CPI}[CPI]{coherent processing interval}
\acro{EOT}[EOT]{extended object tracking}
\acro{rsme}[RSME]{Root-Mean-Square Error}
\acro{ula}[ULA]{uniform linear antenna}
\acro{RCS}[RCS]{Radar Cross Section}
\acro{cnn}[CNN]{convolutional neural network}
\acro{gnn}[GNN]{Graph Neural Network}
\acro{dbf}[DBF]{digital beamforming}
\acro{if}[IF]{intermediate frequency}
\acro{ula}[ULA]{uniform linear array}
\acro{fft}[FFT]{Fast Fourier Transformation}
\acro{cfar}[CFAR]{Constant False Alarm Rate}
\acro{snr}[SNR]{Signal-to-Noise Ratio}
\acro{rmse}[RMSE]{Root Mean Square Error}
\acro{std}[STD]{Standard Deviation}
\acro{lstm}[LSTM]{long short-term memory}
\acro{ol}[OL]{overlap}
\acro{IMM-EOT}[IMM-EOT]{Interacting Multiple Model with Extended Object Tracking}
\acro{RMM-EOT}[RMM-EOT]{Random Matrix Method with Extended Object Tracking}
\acro{HTG}[HTG]{Hierarchical Truncated Gaussian}
\acro{em}[ST]{Standard Tracking}
\acro{immeot}[IMM-EOT]{interacting multiple model with extended object tracking}
\acro{fn}[FN]{False-Negatives}
\acro{LSTM} [LSTM] {Long Short-Term Memory Networks}
\acro{BEV} [BEV] {Bird's-Eye-View}
\acro{MLP} [MLP] {Multilayer Perceptron}
\acro{CS} [CS] {Chirp-Sequence}
\acro{cso} [cso] {coordinate system origin}
\acro{SE} [SE] {Structuring Element}
\acro{PoI} [PoI] {pixel of interest}
\acro{E-RPN}[E-RPN]{Euler-Region-Proposal}
\acro{CSP-Darknet53}[CSP-Darknet53]{Cross-Stage Partial Networks with Darknet53}
\acro{PaNet}[PANet]{Path Aggregation Network}
\acro{YOLO}[YOLO]{You Only Look Once}
\acro{SPP}[SPP]{Spatial Pyramid Pooling}
\acro{md}[\textmu{D}]{micro-Doppler}
\acro{RMM}[RMM]{Random Matrix Method}
\acro{IMM}[IMM]{Interacting Multiple Model}
\acro{CoG}[CoG]{Center of Gravity}
\acro{CuT}[CuT]{Cell under Test}
\acro{roi}[RoI]{Region of Interest}
\acro{adam}[ADAM]{Adaptive Moment Estimation}
\acro{ai}[AI]{artificial intelligence}
\acro{sprb}[SPRB]{set of points on a rigid body}
\acro{WHO}[WHO]{World Health Organization}
\acro{ai}[AI]{artificial intelligence}
\acro{obb}[OBB]{oriented bounding box}
\acro{po}[PO]{Physical Optics}
\acro{rcs}[RCS]{Radar Cross Section}
\acro{los}[LOS]{line-of-sight}
\acro{vru}[VRU]{vulnerable road user}
\acro{sbr}[SBR]{Shooting-and-Bouncing-Ray}
\acro{dnn}[DNN]{Deep Neural Network}
\acro{xai}[XAI]{Explainable Artificial Intelligence}
\acro{cam}[CAM]{class activation maps}
\acro{gan}[GAN]{Generative Adversarial Network}
\acro{piml}[PIML]{Physics-Informed Machine Learning}
\acro{mae}[MAE]{Mean Absolute Error}
\acro{kde}[KDE]{Kernel Density Estimation}
\acro{aeb}[AEB]{Autonomous Emergency Braking}
\acro{iou}[IoU]{Intersection over Union}
\acro{mimo}[MIMO]{Multiple-Input Multiple-Output}
\end{acronym}

\title{Physics-Based Plausibility Assessment of AI-Predicted Vehicle Parameters for Automotive Radar in Pre-Crash Scenarios}

\author{\vspace{-1mm}Rainer Engert\raisebox{1mm}{\orcidlink{0009-0008-6058-7540}}, Patrick Zaumseil\raisebox{1mm}{\orcidlink{0009-0000-9981-8301}} and Thomas Brandmeier\raisebox{1mm}{\orcidlink{0009-0006-1294-3905}}

\thanks{This work has been submitted to the IEEE for possible publication. Copyright may be transferred without notice, after which this version may no longer be accessible.
Received DD Month JJJJ; revised DD Month JJJJ; accepted DD Month JJJJ. Date of publication DD Month JJJJ; date of current version DD Month JJJJ. This work was supported by the German Federal Ministry for Economic Affairs and Energy through the Transformation Hub Accelerate Market Introduction of Autonomous Mobility (MIAMy) under Grant 16THB0010D. (Corresponding authors: Rainer Engert; Patrick Zaumseil)
}

\thanks{
Rainer Engert, Patrick Zaumseil and Thomas Brandmeier are with the Center of Automotive Research on Integrated Safety Systems and Measurement Area (CARISSMA), Technische Hochschule Ingolstadt, 85049 Ingolstadt, Germany (e-mail: rainer.engert@carissma.eu).
}}
\maketitle
{}

\begin{abstract}
Automotive radar is a key sensing modality for safety-critical vehicle functions due to its robustness under adverse weather and lighting conditions. Recent advances in \ac{ai}-based radar perception enable the direct estimation of object parameters from a single radar measurement cycle. However, the internal decision process of data-driven models is typically not explicitly linked to the underlying radar sensing physics. Consequently, object estimates may achieve high statistical accuracy while remaining physically inconsistent with the measured radar signature, particularly in previously unseen scenarios. Such inconsistencies are especially critical for heading estimation, as they directly affect trajectory prediction and collision assessment in pre-crash scenarios.
This paper presents a physics-based consistency assessment framework for validating \ac{ai}-derived vehicle hypotheses in automotive radar perception. The proposed approach generates an expected radar signature from \ac{ai}-estimated object parameters using a physics-based reference model that incorporates occlusion-dependent edge visibility, aspect- and angle-dependent \ac{rcs} modeling, and characteristic wheel-induced \ac{md} signatures. The expected signature is compared with the observed radar point cloud using a Wasserstein-distance-based consistency metric computed from three complementary features: the spatial distribution of detections along object edges, the local detection count, and the radar reflection intensity. In this way, physically implausible object hypotheses can be identified and corrected independently of the underlying neural network architecture and without model retraining or architectural modifications. The proposed consistency module is integrated as a cascaded post-processing stage for vehicle heading estimation. Experimental evaluation on \textbf{\num[detect-weight,detect-family]{1192}} real automotive radar frames demonstrates a statistically significant improvement of approximately {\bfseries\boldmath\SI{19}{\percent}} in heading accuracy compared to the \ac{ai}-only baseline. The results indicate that physics-based consistency assessment of \ac{ai}-derived object hypotheses can effectively reduce large heading errors and improve the reliability and robustness of radar perception systems in safety-critical automotive applications.
\end{abstract}


\section{Introduction}
\label{Intro}

\IEEEpeerreviewmaketitle
\IEEEPARstart{T}{he} ongoing integration of driver assistance and automated driving functions is expected to contribute to an increasingly heterogeneous traffic environment. As a result, vehicles with different levels of automation will operate alongside manually driven vehicles and other traffic participants, leading to more complex interaction patterns and dynamic traffic situations~\cite{Engert_Fahrzeugsicherheit}. Consequently, vehicle safety systems increasingly depend on a reliable perception of surrounding traffic participants and their motion patterns. This is particularly important for traffic safety and pre-crash applications, which impose stringent end-to-end latency requirements while requiring highly dependable prediction and decision-making capabilities~\cite{paden_motion_planning,koopman_vehicle_safety,moubayed_V2X_Computing}. Errors in perception can propagate to downstream prediction and planning modules, ultimately degrading the safety performance of the overall driving systems~\cite{sadat_perceive} and can compromise collision assessment and the activation of pre-crash occupant restraint systems~\cite{pietzsch_pre_crash}

Among the sensing modalities used in automotive applications, radar has gained prominence due to its robustness in adverse weather and lighting conditions and its ability to provide direct measurements of object distance and relative radial velocity~\cite{Hasch_millimeter,Engels_Advances_in_Automotive_Radar,sezgin_Safe_AD}. In particular, radar measurements can contain distinctive \ac{md} signatures that capture object-specific motion patterns and provide additional information beyond object-related parameters such as position, velocity, orientation, and spatial extent~\cite{kamann_Extended_Object_Tracking,vriesman_Rain_Interference}. Among these parameters, the heading angle is particularly important, as it describes the object's orientation and forms part of the vehicle state representations used for state estimation and trajectory planning~\cite{levre_motion_prediction,xin_AV_Heading}. An erroneous heading estimate can adversely affect subsequent trajectory prediction and collision assessment algorithms, thereby impairing the performance of downstream pre-crash vehicle safety systems~\cite{cho_pre_crash,lugner_eval_sensor}. Consequently, automated driving functions rely on continuously available and reliable object information, making accurate object state estimation safety-critical~\cite{feng_Multi_Modal_Detection}. 

To fully exploit the rich information provided by high-resolution radar sensors, recent research has increasingly focused on data-driven radar perception approaches. Deep learning methods have demonstrated promising performance for tasks such as object detection, classification, and oriented bounding box prediction directly from radar data~\cite{zeller_Radar_Vel_Transformer, dong_prob}. In contrast to conventional model-based signal processing pipelines, these approaches learn task-specific feature representations from data and can capture complex relationships that are difficult to model 
analytically. Their growing adoption in automotive perception pipelines, however, raises important questions regarding interpretability and functional safety~\cite{salay_ISO26262, rudin_BB}.

One major challenge is the substantial variability and complexity of real-world traffic scenarios, including rare and previously unseen situations, which cannot be exhaustively captured during development and validation~\cite{Zhang_finding}. Even with comprehensive training data, previously unseen variations in sensing conditions, environmental conditions, and traffic scenarios may arise during deployment, potentially degrading model performance in ways not captured by offline evaluation metrics~\cite{filos_ad,Gopalan_domain}. This issue is particularly relevant for radar sensing, where measurements are generated through electromagnetic wave propagation and are influenced by a variety of physical effects, including multipath propagation~\cite{kamann_Multipath}, target-dependent reflection behavior~\cite{patole_Radar_Review}, coherent interference effects~\cite{signal_Richards_05}, partial occlusions, and overlapping object responses. As a result, the measured radar response of the same target may vary considerably with small changes in aspect angle, target distance, and other influencing factors, posing challenges for robust radar perception~\cite{Hasch_millimeter, Holder_measurement}. This motivates an additional assessment of whether \ac{ai}-estimated object parameters, particularly the heading angle, are physically consistent with the radar response generated by the corresponding object.

To address this, the present work introduces a physics-based plausibility module implemented as a post-processing stage downstream of the object estimation network. The module interprets \ac{ai}-derived object parameters as physical hypotheses and validates them against radar-specific characteristics of the measured point cloud, providing an additional supervisory layer for data-driven perception systems. Physically inconsistent estimates are identified and corrected before they propagate to downstream processing stages, thereby enhancing the reliability of subsequent vehicle safety functions and improving the robustness of \ac{ai}-based radar estimation in safety-critical automotive scenarios.

\subsection{Related Works}
\label{sec:related Works}

The proposed method evaluates whether \ac{ai}-estimated object parameters are physically consistent with the observed radar response. This requires both a model of how object properties are reflected in radar measurements and a framework for assessing the resulting signatures for plausibility. The following section reviews related work from physics-based radar signature modeling and \ac{ai} model validation and discusses why existing approaches do not directly address the problem of runtime plausibility assessment.

\subsubsection{Physics-Based Automotive Radar Signature Modeling}
Automotive radar simulation models differ substantially in their level of abstraction, computational complexity, and electromagnetic fidelity, ranging from simplified object-level sensor models to high-fidelity ray-tracing and full electromagnetic propagation models. Magosi et al.~\cite{magosi_survey_modelling} provide a systematic classification of automotive radar sensor models and show that different validation tasks require different levels of model abstraction.

Early approaches approximate vehicle geometries using virtual scattering centers extracted from ray-tracing simulations~\cite{schuler_ray_tracing}, thereby enabling efficient approximation of their scattering characteristics. The virtual scattering centers are generated by clustering ray-tracing-derived scattering centers using a greedy algorithm rather than by explicitly modeling individual vehicle components.

Semi-physical models extend this by incorporating sensor-specific effects such as range, velocity, angle, detection probability, and \ac{rcs}-dependent reflection behavior. Guo et al.~\cite{guo_radar_modeling} 
and Li et al.~\cite{li_simulation_Model_Radar} combine geometric object representations with sensor-specific uncertainty models, while Magosi and Eichberger~\cite{magosi_simulation_Radar} propose a modular, component-based model with dedicated wheel components to increase simulation realism.

High-fidelity approaches further increase physical realism through ray tracing, \ac{sbr} techniques, or full-scene propagation models. Chipengo et al.~\cite{Chipengo_simulation128, Chipengo_simulation512} generate synthetic radar responses of complete traffic scenes using \ac{sbr}, explicitly distinguishing their approach from simplified point-scatterer models that inadequately capture wave propagation and multipath interactions. Thieling et al.~\cite{Thieling_scalable_radar_sim} propose a scalable GPU-based ray-tracing architecture for dynamic \num{3}D scenes, while Schüßler~\cite{Schuessler_radar_ray_tracing} 
demonstrates an \ac{sbr}-based simulator for high-resolution automotive MIMO radars. Haider et al.~\cite{haider_simulation_fmcw} present a ray-tracing-based \ac{fmcw} radar sensor model that incorporates \ac{mimo} signal processing and sensor-specific non-idealities, such as phase noise and mixer nonlinearity. More generally, Magosi~\cite{magosi_eval_radar} emphasizes that the architecture and capabilities of radar sensor models should be selected according to the respective development phase.

Several studies specifically investigate \ac{md} signatures arising from rotating wheels and moving vehicle components, demonstrating that radar observations are determined not only by object position and extent but also by aspect angle, component visibility, scattering characteristics, and motion-induced Doppler effects~\cite{Ram_simulation_md_human, Chipengo_physic_radar, wald_Ray_tracing_MD, kamann_Extended_Object_Tracking, Zaumseil_Radar_Signature}. In the automotive context, component-oriented sensor models incorporate dedicated wheel models with associated \ac{md} signatures to reproduce wheel-specific radar phenomena~\cite{magosi_simulation_Radar}.

However, most existing radar simulation approaches are designed for sensor validation, synthetic data generation, or target classification, and they generate radar measurements from known simulated scenes rather than evaluating predicted object hypotheses against observed data. This distinction motivates the second research area reviewed below.

\subsubsection{AI Model Validation in Radar Perception}
Despite the growing deployment of \ac{ai}-based perception in autonomous systems, runtime monitoring and safety assurance of \ac{dnn}-based perception remain active challenges \cite{huang_safey_dnn}. Due to their black-box nature, \acp{dnn} often lack transparency, raising concerns regarding reliability, trustworthiness, and safety, particularly in safety-critical applications such as autonomous driving~\cite{Guidotti_BB}. Prior research has examined the vulnerability of deep learning models to adversarially perturbed inputs~\cite{Ian_Explaining_and_Harnessing_Adversarial_Examples, Szegedy_Intriguing_properties_of_neural_networks}, and robustness benchmarks have been proposed to evaluate neural networks under realistic corruptions and perturbations~\cite{Hendrycks_Benchmarking_Neural_Network_Robustness_to_Common_Corruptions_and_Perturbations}.

In radar-based perception, adverse weather conditions, multipath reflections, and ghost objects can significantly degrade data quality and reduce the \ac{snr}~\cite{Infuence_Rain, kamann_Multipath}, while fluctuations in \ac{rcs} may further destabilize object detection. These sensor-specific effects motivate the use of \ac{xai} methods to improve the interpretability of deep learning models and to better understand the decision-making process and validate model predictions~\cite{Ribeiro_why, Petsiuk_RISE, Ramprasaath_grad_cam}.

Beyond performance-based evaluation, recent research emphasizes the importance of validating machine learning models against physical principles and domain knowledge. Furthermore, physics-informed machine learning models incorporate physical constraints and prior knowledge into the learning process to promote physically consistent, plausible, and trustworthy model predictions~\cite{naser_validate, Nghiem_physics_informed, Meng_Physics_ML}. In autonomous driving, recent work has shown that conventional accuracy metrics alone are insufficient to assess trajectory prediction quality, as learning-based models may generate physically infeasible trajectories. Consequently, physically infeasible metrics have been introduced as a complementary evaluation criterion to account for physically infeasible outputs from learning-based models~\cite {Chengyue_WAKE}.

More recently, runtime monitoring and validation of \ac{ai}-based perception systems have emerged as an active research direction. Rahman et al.~\cite{Rahman_run_time_monitoring} survey runtime monitoring approaches for machine learning in robotic perception, arguing that design-time verification alone is insufficient for unknown deployment environments and that continuous runtime assessment of reliability is necessary. Keser et al.~\cite{Keser_verification_Object_detectors} propose an interpretable, model-agnostic plausibility verification framework for \num{2}D object detectors based on domain-invariant concept bottleneck models, enabling operation-time plausibility checks without requiring access to the detector's internal representations. Aslam et al.~\cite{Aslam_runtime_validation} propose a function monitor for the runtime validation of camera- and LiDAR-based \ac{ai} perception, addressing the lack of complete requirements specifications assumed by ISO~\num{26262} and ISO~\num{21448}.

Existing validation approaches, however, are predominantly applied offline against labeled datasets and do not enable frame-level plausibility evaluation of individual object hypotheses at runtime. The present work addresses this gap by using physics-based radar modeling as a hypothesis-conditioned reference. The \ac{ai}-predicted object position, dimensions, and heading are used to generate an expected radar signature, which is then compared with the observed radar point cloud via Wasserstein distances to assess the physical consistency of the prediction at runtime. Unlike existing simulation approaches, which generate measurements from known scenes, the proposed method inverts this process: it evaluates how well a predicted object hypothesis explains the observed sensor data.

\subsection{Objectives and Contributions}
\label{sec:research}
The works discussed above demonstrate that physics-based radar modeling and \ac{md} signature simulation have been extensively studied for radar sensor validation, data augmentation, and target classification. However, several key distinctions separate the proposed approach from existing methods.

Existing radar simulation approaches generate radar measurements from known, fully specified scenes and are designed for sensor validation, synthetic data generation, or target classification. None of the reviewed approaches use a predicted object hypothesis to construct an expected radar signature and evaluate it against an observed point cloud at runtime. The direction of inference is inverted in the proposed method: rather than simulating what a sensor would measure given a known scene, the proposed framework assesses how well an \ac{ai}-predicted object hypothesis explains the observed radar data.

Existing \ac{ai} validation approaches in radar perception are predominantly applied offline, relying on labeled datasets and statistical performance metrics such as \ac{iou}, precision, and recall, evaluated against ground-truth annotations. While these metrics provide a global assessment of model quality, they do not support frame-level plausibility evaluation of individual object hypotheses at runtime and are not applicable in deployment scenarios without ground truth.

Physics-informed machine learning and \ac{xai} methods address model transparency and physical plausibility at the level of model internals, using gradients, feature activations, or physical constraints during training. These approaches operate at the model level and do not perform signal-level consistency checks between predicted object parameters and observed radar measurements during inference.

The proposed method addresses all three gaps simultaneously. An expected radar signature is constructed directly from the \ac{ai}-predicted object hypothesis, including position, dimensions, and heading angle, using a physics-based reference model that accounts for occlusion-dependent edge visibility, aspect- and angle-dependent \ac{rcs} variations, characteristic wheel-induced \ac{md} signatures, and the resulting spatial distribution, intensity, and density of radar detections. The expected signature is then compared with the observed radar point cloud using a Wasserstein-distance-based score to perform a frame-level physical-consistency assessment, independently of the \ac{ai} model's internal structure and without requiring retraining or architectural modifications.

The primary objective of this work is therefore to introduce a physics-based consistency assessment framework for validating \ac{ai}-derived vehicle hypotheses in automotive radar perception. Its main contributions are:
\begin{itemize}
    \item A physics-based radar signature generation model that constructs expected radar signatures from \ac{ai}-derived vehicle object parameters, thereby enabling the physical validation of individual object hypotheses from a single radar observation. The model combines geometric visibility constraints, aspect-dependent scattering characteristics, velocity-dependent Doppler effects, and detection-density modeling to represent the expected spatial, intensity, velocity, and detection-count distributions of radar observations.

    \item A neural-network-independent physical consistency assessment that compares expected and observed radar signatures using a Wasserstein-distance-based metric. In contrast to conventional prediction confidence measures, the proposed method evaluates whether \ac{ai}-derived object hypotheses are physically supported by the measured radar response and identifies physically implausible object estimates.

    \item A cascaded post-processing framework for radar-based vehicle estimation that refines \ac{ai}-derived heading estimates without retraining the model or modifying the architecture. Experimental evaluation on real automotive radar data demonstrates a statistically significant improvement of approximately \SI{19}{\percent} compared to the \ac{ai}-only baseline.
\end{itemize}

\subsection{Outline}
The remainder of this paper is organized as follows: Section~\ref{sec:methods} describes the two-stage radar-based object detection pipeline, covering the radar signal processing chain and the adopted ComplexYOLO detector. Section~\ref{sec:model} presents the physics-based assessment framework, including the geometric data association, the physical model, and the rotation-based heading correction. Section~\ref{sec:measure} describes the experimental setup, including the radar sensor configuration, the empirical model parameters, and the validation and test datasets used for evaluation. Section~\ref{sec:results} reports the quantitative and qualitative evaluation of the proposed approach on both validation and test data. Section~\ref{sec:discussion} interprets the results in the broader context of radar-based vehicle perception and discusses the identified limitations. Finally, Section~\ref{sec:conclusion} summarizes the key findings and outlines directions for future work.

\section{Radar-Based Object Detection}
\label{sec:methods}

This section describes the two-stage processing pipeline that transforms raw radar measurements into estimates of object parameters. First, the raw radar data is processed using a signal-processing chain to extract a structured point cloud representation. Second, the resulting representation is passed to an \ac{ai}-based object detector, which regresses the object parameters required by the subsequent plausibility module.

\subsection{Signal Processing}
\label{sec:signal}
Modern automotive radar sensors operate according to the chirp sequence \ac{fmcw} principle by transmitting rapidly frequency-swept chirps \cite{Reference:Xu_FMCW_Chirps}. Within a single chirp, the resulting beat frequency primarily encodes the target range, while the Doppler-induced frequency shift is negligible due to the short chirp duration. Consequently, the radial velocity $v_d$ is not estimated from an instantaneous frequency offset. Instead, it is determined from the phase progression between consecutive chirps, which is solely caused by the target motion \cite{costa1984signal, Stove1992LinearFR}.

To estimate the azimuth angle, \ac{dbf} is applied. The directional information is contained in the spatial phase progression across the receive antennas of the \ac{ula}. A widely adopted approach for joint range, Doppler, and azimuth processing is to compute a three-dimensional \ac{fft}, yielding spectral components along all three dimensions \cite{Engels_Advances_in_Automotive_Radar, Diss_Heidenreich}. The corresponding discrete spectral formulation is expressed as
\begin{align}
    Y(\nu, \mu, \rho) = \sum_{l_s = 0}^{L_s - 1} \sum_{m_s = 0}^{M_s - 1} \sum_{n_s = 0}^{N_s - 1} w_\nu(l_s) w_\mu(m_s) w_\rho(n_s) \notag \\ \cdot x(l_s, m_s, n_s) \cdot e^{-j(\nu l_s + \mu m_s + \rho n_s)}.
    \label{eq:ffts}
\end{align}
The signal $x(l_s, m_s, n_s)$ denotes the sampled intermediate-frequency signal arranged in a three-dimensional data cube. The indices $l_s, m_s$, and $n_s$ correspond to range samples, chirp index, and antenna element index with respective sizes $L_s$,  $M_s$, and $N_s$. The functions $w_\nu(l_s)$, $w_\mu(m_s)$, and $w_\rho(n_s)$ represent normalized window functions applied along the respective dimensions to mitigate spectral leakage. The discrete frequency variables $\nu = (2\pi/L_s) \cdot l_s$, $\mu = (2\pi/M_s) \cdot m_s,$ and $\rho = (2\pi/N_s)\cdot n_s$ define the spectral bins associated with range, radial velocity, and azimuth angle. The mapping to physical parameters is obtained via radar system characteristics such as chirp slope, carrier frequency, sampling frequency, and antenna spacing \cite{barrick1973}. In contrast to conventional \ac{fft}-based angular spectrum estimation, this work employs the Burg algorithm \cite{Reference:Burg}. As a parametric autoregressive spectral estimator, the Burg method provides enhanced angular resolution compared to windowed periodograms and exhibits reduced spectral leakage, enabling improved separation of closely spaced angular components \cite{Reference:Burg2}.

For power detection, the magnitudes of the complex range-Doppler spectra are computed across the $N_s$ antennas and combined using non-coherent integration \cite{signal_Richards_05}:
\begin{align}
    \abs{\overline{X}(\nu,\mu)}_{N_s} = \frac{1}{N_s} \sum^{N_s}_{n_s=1} \abs{X(\nu,\mu)}_{n_s}
    \label{eq:power_detection}
\end{align}

This integration increases the effective signal-to-noise ratio and improves the separability of target reflections from noise. Based on the integrated range-Doppler spectrum, a first \ac{cfar} detection algorithm is performed to extract significant detections with adaptive thresholding \cite{rohling}. The resulting detections are grouped using \ac{dbscan} clustering in order to suppress isolated false alarms and consolidate spatially coherent reflections \cite{esterDbscan}. To further refine the remaining detections, the \ac{cfar} procedure is repeated on the pruned range-Doppler representation, followed by \ac{dbscan} clustering again. This iterative detection-clustering procedure reduces the residual clutter response and stabilizes target hypotheses prior to high-resolution azimuth estimation.

\subsection{Adopted ComplexYOLO Radar Detector}
\label{sec:ki-model}
The processed radar point cloud is subsequently passed to an \ac{ai}-based object detector to regress vehicle bounding boxes and associated object parameters. In this work, the radar-based object detection model introduced in our previous publication \cite{zaumseil_YOLO} is adopted as a fixed framework. The model builds upon the anchor-based ComplexYOLO architecture originally developed for LiDAR data \cite{Simon_ComplexYOLO} and has been specifically adapted for high-resolution automotive radar data and single-frame pre-crash object detection. All architectural components, radar-specific \ac{BEV} representations, \ac{md} feature integration, training procedures, and dataset descriptions are provided in detail in \cite{zaumseil_YOLO} and are therefore not repeated here.

For each radar measurement cycle, the detector processes the radar-derived \ac{BEV} representation and outputs a set of object parameters. Each parameter set is represented by a regressed kinematic state vector containing the object's center position, spatial dimensions, heading angle, and estimated target velocity. The \ac{ai} model operates frame-wise and does not rely on temporal tracking or multi-frame aggregation. In the present work, the detector serves as a hypothesis generator, with its outputs serving as input to the proposed plausibility module. The complete parameter set, excluding the target velocity, is forwarded per radar frame to the subsequent plausibilization stage without architectural modifications or additional temporal filtering. Based on the regressed geometric object parameters, a radar-consistent measurement model is instantiated to generate expected radar detections, which are then compared with the actually enclosed radar point cloud to assess physical consistency. This design choice reflects the targeted pre-crash application, where the sudden appearance of an object in the near-field may not allow for object-tracking convergence and requires immediate frame-based evaluation.
\section{Physics-Based Assessment Framework}
\label{sec:model}

\begin{figure*}[t]
    \centering
    \includegraphics[width=1\linewidth]{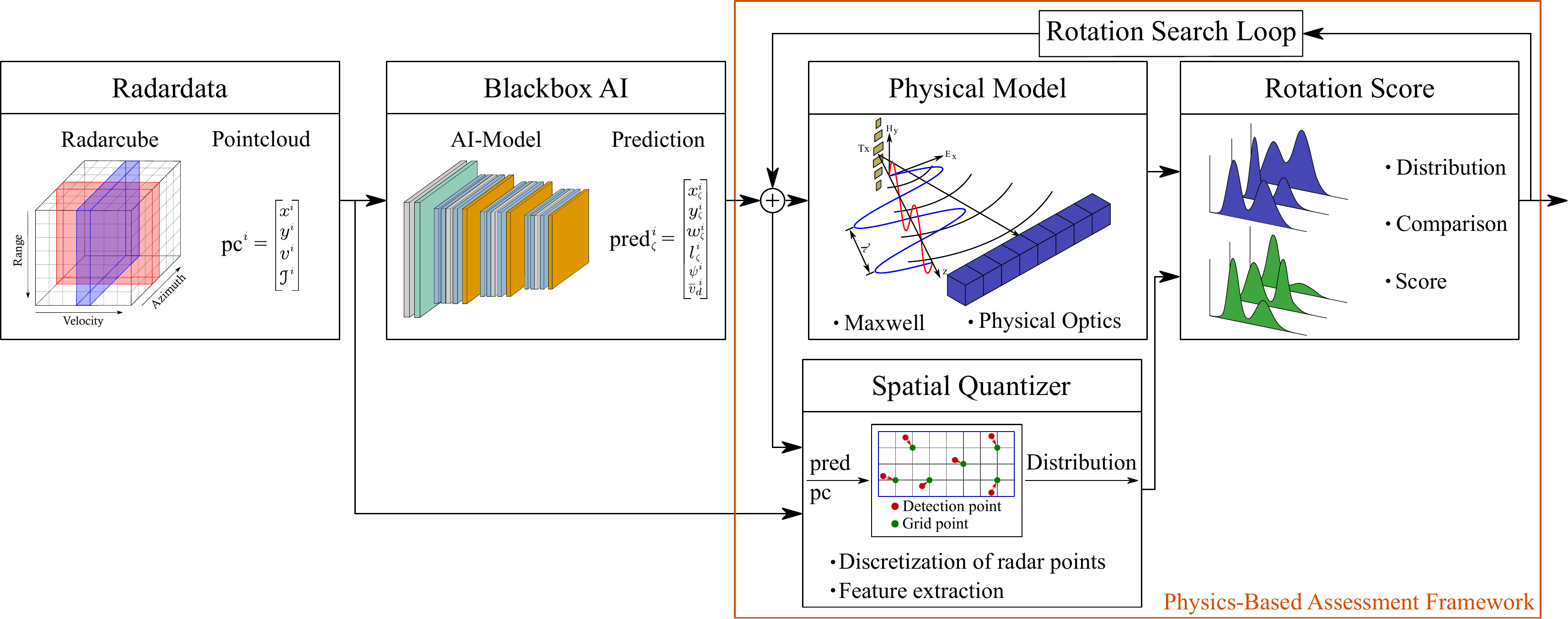}
    \caption{Overview of the proposed pipeline. The \ac{ai} model processes the radar point cloud and outputs an initial vehicle box prediction, which is forwarded to the physical model and the spatial quantizer. The spatial quantizer discretizes the observed radar point cloud into reference distributions, which are compared against the expected distributions computed by the physical model in the rotation score block. Within the \textit{Physics-Based Assessment Framework} (orange box), the predicted orientation is iteratively rotated and re-evaluated until all rotations are scored. The candidate with the minimum score is selected as the optimal yaw correction.}
    \label{fig:model_archi}
\end{figure*}

The entire proposed pipeline is outlined in Fig.~\ref{fig:model_archi}. Radar point cloud data, together with an initial bounding box prediction obtained from a black-box \ac{ai} model, serve as the primary inputs. The pipeline consists of two parallel stages followed by an iterative search loop, each described in detail in the following subsections. The spatial quantizer decomposes the predicted bounding box and associates radar detections with a bin grid. Simultaneously, the physical model computes the expected distributions of radar signatures for the predicted bounding box based on the \ac{ai}-predicted object state. Within the rotation search loop, the bounding box orientation is iteratively rotated over $\theta \in [\theta_{\text{min}}, \theta_{\text{max}}]$. For each candidate $\theta$, the physical model and the spatial quantizer recompute the radar signature distributions, which are then compared in the rotation score block. The candidate yielding the minimum aggregated score is selected as the optimal yaw correction.

\subsection{Spatial Quantizer}
\label{sec:data_asso}
The spatial quantizer serves as the interface between raw radar detections and the estimates produced by a precursor \ac{ai} model. This stage uses the \ac{obb} as a dynamic reference frame to discretize the radar point cloud into structured reference distributions on a per-frame basis for subsequent comparison with the physical model.

\subsubsection{Spatial Discretization}
To convert the radar point cloud into a structured and comparable representation, a spatial discretization based on the \ac{ai}-predicted object state is applied. Given the predicted state vector of the \ac{obb}, comprising the center point $(x_M, y_M)$, the object dimensions (width $w$, length $l$), and the yaw angle $\psi^{\text{AI}}$, the interior of the \ac{obb} is partitioned into a fixed grid of \SI{10}{\centi\meter} $\times$ \SI{10}{\centi\meter} bins. Within the rotation search loop, $\psi^{\text{AI}}$ is iteratively modified by the candidate rotation $\theta$, such that the bin grid is recomputed for each candidate orientation. A radar detection is associated with a specific bin $B_{i,j}$ using a point-in-polygon test. The occupancy of each bin is defined as the number of detections $N_{i,j}$ contained within it, while the geometric center of each bin serves as the spatial coordinate for the associated radar detections.

\begin{figure}[b!]
    \centering
    \includegraphics[width=0.9\linewidth]{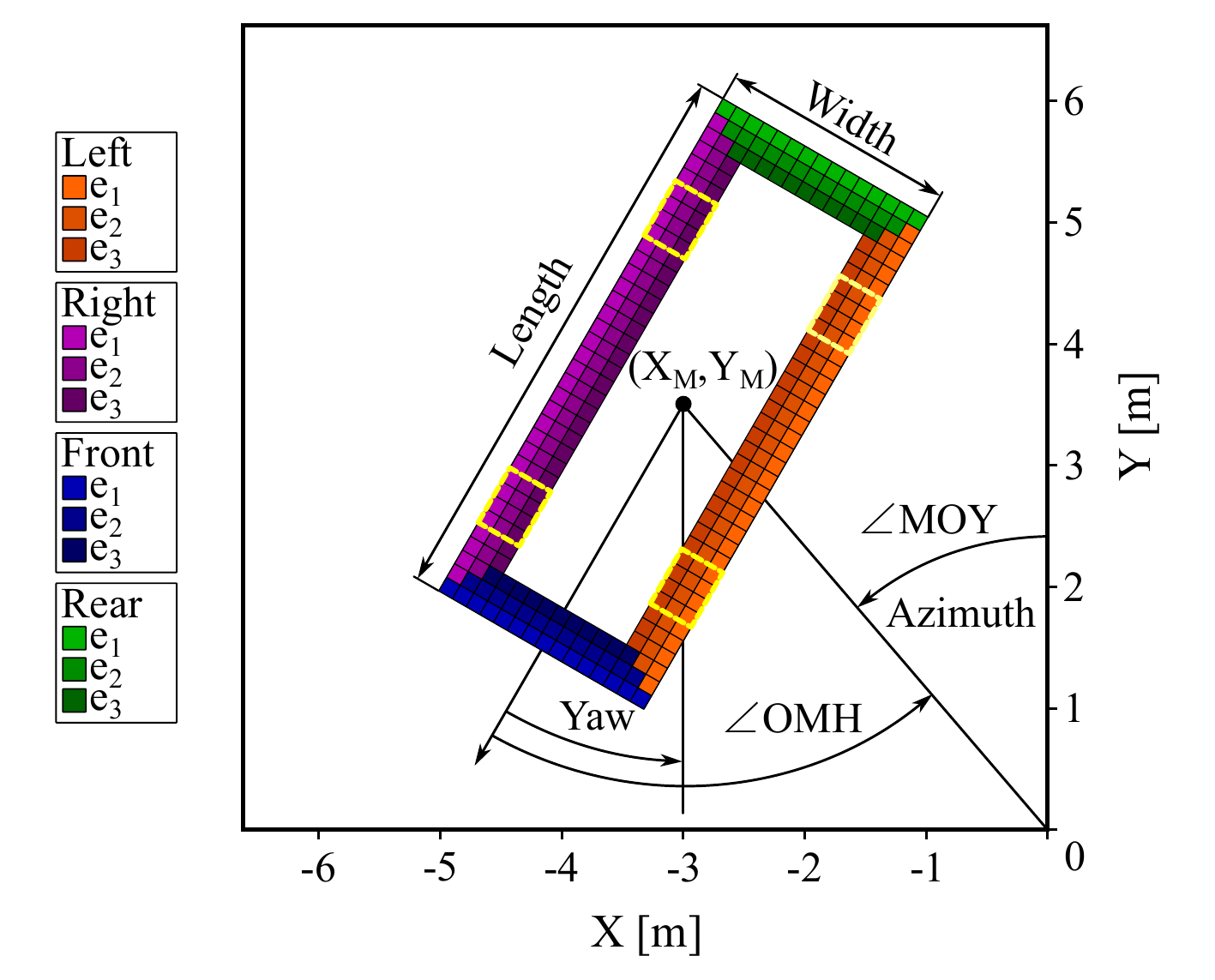}
    \caption{Illustration of the bounding box edge decomposition based on the \ac{ai}-predicted object state $(X_M, Y_M, \text{length}, \text{width}, \psi)$. The \ac{obb} interior is discretized into \SI{10}{\centi\meter} $\times$ \SI{10}{\centi\meter} bins, grouped into four directional sides (front, rear, left, right) and three edge zones $e_1$, $e_2$, $e_3$ from the outer boundary towards the interior. Bins associated with wheel regions are indicated by a yellow dashed outline.}
    \label{fig:box_edges}
\end{figure}

\subsubsection{Edge Decomposition and Feature Extraction}
The discretized grid is decomposed into four directional sides (front, rear, left, and right), each further subdivided into three edges $e_1$, $e_2$, and $e_3$ arranged from the outer boundary towards the interior of the object, as illustrated in Fig.~\ref{fig:box_edges}. The choice of three edges with a bin size of \SI{10}{\centi\meter} $\times$ \SI{10}{\centi\meter} corresponds to a total edge depth of \SI{30}{\centi\meter}, reflecting the typical penetration depth of
radar reflections into vehicle structures.

For each edge, three complementary features are extracted. Each radar detection point is first assigned to the nearest bounding box edge and then projected onto that edge, yielding a scalar coordinate indicating the detection's position along the edge. This projection coordinate is subsequently discretized into a fixed number of bins, from which three per-bin quantities are derived. The spatial density of projected detections reflects the distribution of measurements along the edge. The absolute detection count per bin quantifies local measurement density as a function of position. The mean radar reflection intensity per bin encodes physical reflection properties such as material characteristics and angle-of-incidence effects.

Additionally, the Euclidean distance $d_{\text{euc}}$ between each bin centroid $C_{\text{bin}}$ and the mean centroid of all radar detections $C_{\text{radar}}$ is computed as
\begin{align}
    d_{\text{euc}} = \left\lVert C_{\text{bin}} - C_{\text{radar}} \right\rVert_2,
\end{align}
providing a displacement-based reference for subsequent comparison with the physical model. The resulting structured feature set $\mathrm{Output}_{\text{AI}}$ comprises, for each directional side and edge zone, a detection count distribution, an intensity distribution, and a displacement distribution derived from $d_{\text{euc}}$.

\subsection{Physical Model}
\label{sec:physical_model}
The physical model serves as the deterministic backbone of the evaluation framework, providing physics-based estimates of radar signatures for a given \ac{ai}-predicted vehicle state and a known sensor position. Rather than requiring detailed geometric knowledge of the actual vehicle, the model operates on an abstract vehicle representation consisting of the predicted bounding box and associated wheel regions, as illustrated in Fig.~\ref{fig:box_edges}. By synthesizing geometric visibility constraints, electromagnetic scattering characteristics, and kinematic Doppler contributions, the model generates a synthetic detection distribution that serves as a reference for evaluating \ac{ai}-based estimates. The following subsections detail the four components of the physical model in the order in which they are applied.

\subsubsection{Occlusion Handling}
To identify occluded object edges, the visibility of each edge is determined from the \ac{ai} object state, comprising the center point $M^{\text{AI}} = (x_M, y_M)$, the object dimensions, and the yaw angle $\psi^{\text{AI}}$. Throughout this section, $(\cdot)^{\text{AI}}$ denotes quantities directly predicted by the \ac{ai} model.
Given $\psi^{\text{AI}}$, the heading direction vector is defined as
\begin{align}
    \mathbf{d}_{\psi} = [\cos(\psi^{\text{AI}}),\, \sin(\psi^{\text{AI}})].
\end{align}
A heading point $H$ is constructed along this direction as
\begin{align}
    H = M^{\text{AI}} + \kappa \mathbf{d}_{\psi},
\end{align}
where $\kappa$ corresponds to half the bounding box length, scaling the orientation vector to the spatial extent of the detected object. Based on $M^{\text{AI}}$, the sensor origin $O$, and the heading point $H$, two angles are computed,as illustrated in Fig.~\ref{fig:box_edges}: the angle $\angle OMH$ between the sensor origin $O$, the object center $M^{\text{AI}}$, and the heading point $H$, and the angle $\angle MOY$ (Azimuth) between the object center $M^{\text{AI}}$, the sensor origin $O$, and the local object $y$-axis. Together, these angles fully describe the relative orientation between the radar sensor and the predicted object pose.

The visibility of the four object edges (front, rear, left, right) is determined by rule-based geometric conditions. The lateral edges are classified from $\angle OMH$: the \textit{right} edge is visible if
\[
    0^\circ < \angle OMH < 180^\circ,
\]
while the \textit{left} edge is visible in the complementary range. A rear-facing configuration is identified when
\[
    180^\circ < \angle OMH < 360^\circ,
\]
while the remaining orientations correspond to a front-facing configuration. At the boundary cases $\angle OMH \in \{0^\circ, 90^\circ, 180^\circ, 270^\circ\}$, only a single edge is classified as visible to avoid ambiguity: the front edge at \SI{0}{\degree}, the right edge at \SI{90}{\degree}, the rear edge at \SI{180}{\degree}, and the left edge at \SI{270}{\degree}. To further stabilize the front-rear classification near angular transition regions, $\angle MOY$ is additionally evaluated: the front edge is classified as visible only if
\[
    110^\circ \leq \angle MOY \leq 250^\circ,
\]
serving as a secondary criterion that overrides the primary $\angle OMH$ classification exclusively in ambiguous transition regions. In all remaining cases, $\angle OMH$ alone determines the visible edge configuration.

\subsubsection{Intensity Modeling}
The radar intensity assigned to each edge bin is modeled using a simplified scattering formulation inspired by the \ac{po} approximation combined with the monostatic radar range equation. The objective is not to reproduce the full electromagnetic scattering of complex vehicle geometries but to capture the dominant characteristics of radar reflections from elongated structures in a computationally efficient and physically interpretable manner.

Electromagnetic wave propagation in a homogeneous, source-free region is governed by the wave equations for the electric ($\mathbf{E}$) and magnetic ($\mathbf{H}$) fields~\cite{balanis2012advanced}
\begin{align}
    \Delta \mathbf{E} - \mu \varepsilon \frac{\partial^2 \mathbf{E}}
    {\partial t^2} = 0,
    \qquad
    \Delta \mathbf{H} - \mu \varepsilon \frac{\partial^2 \mathbf{H}}
    {\partial t^2} = 0.
\end{align}
Assuming time-harmonic fields of the form $e^{j\omega t}$, the wave equations reduce to the Helmholtz equation~\cite{balanis2012advanced}
\begin{align}
    \nabla^2 \mathbf{E} + k^2 \mathbf{E} = 0,
    \qquad
    \nabla^2 \mathbf{H} + k^2 \mathbf{H} = 0,
\end{align}
with wavenumber $k = 2\pi/\lambda$, where $\lambda$ denotes the radar wavelength.

To model the scattering from vehicle surfaces, the \ac{po} approximation is applied. When a radar wave with incident magnetic field $\mathbf{H}_{\text{inc}}$ illuminates a conducting surface, the induced surface current density on the illuminated side is approximated by~\cite{balanis2012advanced, harrington2001time}
\begin{align}
    \mathbf{J} \approx 2\,\hat{\mathbf{n}} \times \mathbf{H}_{\mathrm{inc}},
\end{align}
where $\hat{\mathbf{n}}$ denotes the outward surface normal. These induced currents act as secondary sources of radiation. The resulting scattered electric field received at the radar is obtained by inserting $\mathbf{J}$ into the radiation integral~\cite{balanis2012advanced}
\begin{align}
    E_{\mathrm{RX}} =
    -\frac{jk}{4\pi}
    \int_{S}
    (\mathbf{J} \times \mathbf{r})
    \frac{e^{-jkR}}{R}
    \, \mathrm{d}S,
\end{align}
where $S$ denotes the illuminated surface, $R$ the distance between the radar and the scattering point, and $\mathbf{r}$ the unit vector pointing from the scattering point towards the radar. For a uniform current distribution along a finite edge segment of effective length $L_{\mathrm{eff}}$, the far-field angular response follows \cite{knott2004radar}
\begin{align}
    E_{\mathrm{sc}}(\theta) \propto
    \mathrm{sinc}\!\left(k L_{\mathrm{eff}} \sin(\theta)\right),
\end{align}
where $\theta$ denotes the aspect angle between the surface normal and the radar line-of-sight. To account for geometric visibility effects, the effective edge length $L_{\text{eff},i}$ represents the portion of the edge segment that contributes to radar scattering at the aspect angle $\theta_i$. It is computed as
\begin{align}
    L_{\mathrm{eff},i} =
    \mathrm{clip}\!\left(L |\cos(\theta_i)|,\, L_{\min},\, L\right),
\end{align}
where $L$ denotes the full segment length and $L_{\min}$ a minimum threshold to account for residual scattering at grazing angles.

Since radar power is proportional to the squared magnitude of the scattered field, the power angular response follows directly from $|E_{\mathrm{sc}}|^2 \propto \mathrm{sinc}^2\!\left(k L_{\mathrm{eff},i} \sin(\theta_i)\right)$. To account for residual non-directional scattering contributions at grazing angles arising from surface roughness and higher-order reflections not captured by the \ac{po} approximation, a floor term $\varepsilon$ is introduced~\cite{knott2004radar,skolnik2008radar}, yielding the angular gain response
\begin{align}
    g(\theta_i) =
    (1-\varepsilon)
    \,
    \mathrm{sinc}^2\!\left(k L_{\mathrm{eff},i} \sin(\theta_i)\right)
    + \varepsilon,
\end{align}
where $\varepsilon \in [0,1]$ is an empirically chosen parameter, and $(1-\varepsilon)$ ensures that the gain is normalized to unity at broadside incidence.

The relative aspect angle $\theta_i$ describes the angular relationship between the vehicle surface normal at bin $i$ and the radar line-of-sight. It is computed from the \ac{ai}-predicted yaw angle $\psi^{\text{AI}}$ and the bin azimuth $\alpha_i$ as
\begin{align}
    \theta_i =
    \mathrm{wrap}_{(-180^\circ,180^\circ)}(\psi^{\text{AI}} - \alpha_i),
    \label{eq:theta}
\end{align}
where $\alpha_i$ denotes the azimuth angle from the radar sensor to bin $i$, and the wrap operation ensures that $\theta_i$ remains within $(-180^\circ, 180^\circ]$. This angle serves as the central input to all aspect-dependent computations described below.

The \acf{rcs} is modeled as a function of the aspect angle, assuming planar surface elements whose reflectivity decreases with increasing angle of incidence~\cite{knott2004radar}:
\begin{align}
    \sigma(\theta_i) = \sigma_0 |\cos(\theta_i)|,
\end{align}
with the base \ac{rcs} value $\sigma_0$ depending on the semantic bin assignment:
\begin{align}
    \sigma_0 =
    \begin{cases}
        \Sigma_{\text{wheel}},   & \text{wheel-related bins},\\
        \Sigma_{\text{chassis}}, & \text{chassis-related bins}.
    \end{cases}
\end{align}
Here, $\Sigma_{\text{wheel}}$ and $\Sigma_{\text{chassis}}$ are empirically chosen (section~\ref{sec:empirical}) base \ac{rcs} values for wheel-related and chassis-related bins, respectively, reflecting the higher radar reflectivity of rotating wheel structures compared to flat chassis surfaces.

Combining the angular gain response $g(\theta_i)$, the aspect-dependent \ac{rcs} $\sigma(\theta_i)$, and the monostatic radar range equation~\cite{skolnik2008radar}, the model-predicted intensity at bin position $\mathbf{p}_i$ is obtained as:
\begin{align}
    I_i = P_r(\mathbf{p}_i) =
    \frac{P_t G_t G_r \lambda^2}{(4\pi)^3 R_i^4}
    \,\sigma(\theta_i)\, g(\theta_i),
    \label{eq:intensity}
\end{align}
where $P_t$ denotes the transmitted power, $G_t$ and $G_r$ are the transmit and receive antenna gains, and $R_i = \|\mathbf{p}_i - O\|$ is the range from the radar sensor to bin $i$.

\subsubsection{Velocity Modeling}
A radial velocity is assigned to each bin to represent the anticipated Doppler response. The normalized line-of-sight vector from the radar sensor to bin $i$ is
\begin{align}
    \mathbf{l}_i =
    \frac{\mathbf{p}_i - O}{\|\mathbf{p}_i - O\|},
\end{align}
where $O$ denotes the position of the radar sensor and $\mathbf{p}_i$ the position of bin $i$.

The object motion is described by a longitudinal velocity $v_0$ along the heading direction defined by $\psi^{\text{AI}}$:
\begin{align}
    \mathbf{v} = v_0[\cos \psi^{\text{AI}},\, \sin \psi^{\text{AI}}].
\end{align}
Since radar sensors measure only the velocity component along the line-of-sight, the radial velocity at bin $i$ is obtained as
\begin{align}
    v_i = \mathbf{v} \cdot \mathbf{l}_i = v_0 \cos(\psi^{\text{AI}} -
    \theta_i),
\end{align}
where $\theta_i$ denotes the relative aspect angle between the heading direction and the radar line-of-sight at bin $i$, as defined in \eqref{eq:theta}.

\begin{figure}[t]
    \centering
    \includegraphics[width=1.0\linewidth]{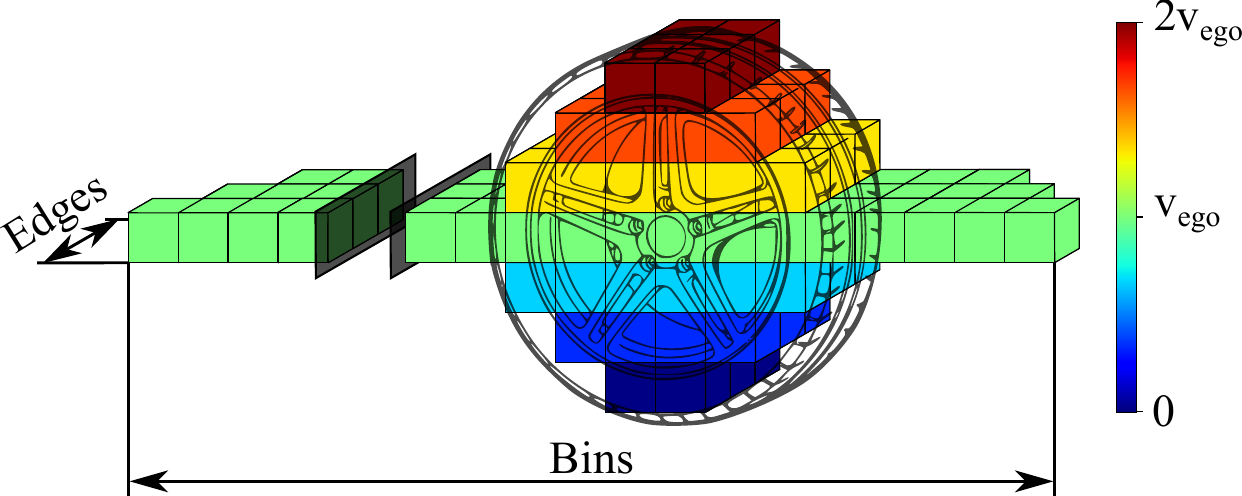}
    \caption{Illustration of the velocity modeling. Due to wheel rotation, different bin locations on the wheel produce distinct radial velocity components.}
    \label{fig:wheel_velocity_bins}
\end{figure}

Chassis bins correspond to stationary surface regions of the vehicle body and receive a single radial velocity $v_i$ derived from the projected object velocity along the radar line-of-sight. Wheel bins, by contrast, correspond to rotating tire regions in which the rotational motion adds an additional velocity component that varies across the tire surface, from zero at the wheel center to twice the translational velocity at the outer rim. This spatially varying contribution yields a range of radial velocities rather than a single value. Consequently, a discretized velocity interval $[v_{\min}, v_{\max}]$ is assigned to each wheel bin to represent the full range of radial velocity contributions arising from wheel rotation, as illustrated in Fig.~\ref{fig:wheel_velocity_bins}. The interval boundaries are determined from the geometric distance between the bin position and the wheel center and are subsequently projected onto the radar line-of-sight. The number of discrete velocity steps covered by the interval is then given by
\begin{align}
    K_i = \left\lfloor \frac{v_{i,\max} - v_{i,\min}}{\Delta v} \right\rfloor,
\end{align}
where $\Delta v$ denotes the velocity resolution. $K_i$ thus represents the number of radial velocity bins to which detections of bin $i$ can be mapped due to wheel rotation.

\subsubsection{Detection Count Modeling}
\label{sec:model_detcount}
The per-bin detection count of the physical model is given by
\begin{align}
    D_i = N_i \cdot I_i^{\mathrm{norm}} \cdot
    w(\theta_i) \cdot f_{\mathrm{FOV}},
    \label{eq:detection_count}
\end{align}
where $N_i$ denotes the nominal detection count, $I_i^{\mathrm{norm}}$ the normalized intensity weighting, $w(\theta_i)$ the angular weighting function, and $f_{\mathrm{FOV}}$ the field-of-view indicator. Each factor captures a distinct physical aspect of the radar response and is derived as follows.

The nominal detection count $N_0(R_i)$ is derived from the sensor resolution and the bin geometry as
\begin{align}
    N_0(R_i) = \frac{A_{\text{bin}}}{\delta_r \cdot R_i \cdot \delta_\phi},
    \label{eq:N0}
\end{align}
where $A_{\text{bin}}$ denotes the bin area, $\delta_r$ the range gate length, $\delta_\phi$ the azimuth angular resolution, and $R_i$ the distance from the sensor to bin $i$. The sensor-specific values are given in Section~\ref{sec:empirical}. For each of the four bounding box sides and their three associated edge zones, the computation distinguishes between chassis bins, which cover the vehicle body surface, and wheel bins, which account for the additional velocity spread introduced by tire rotation. For chassis bins, $N_i = N_0(R_i)$. For wheel bins, the rotational motion contributes additional detections spread across $K_i$ discrete velocity bins, which are added to the chassis contribution:
\begin{align}
    N_i = \begin{cases}
        N_0(R_i) + K_i               & \text{(wheel bin, visible)}, \\
        N_0(R_i) + \dfrac{K_i}{2}   & \text{(wheel bin, occluded)}, \\
        N_0(R_i)                     & \text{(chassis bin)}.
    \end{cases}
\end{align}
For occluded wheel bins, only half of the velocity contributions are retained, accounting for the attenuation of wheel-related reflections by the vehicle body. To avoid boundary artifacts, the first and last bins of each edge are excluded.

The model-predicted intensity $I_i$, as defined in \eqref{eq:intensity}, provides a physically motivated per-bin intensity estimate that reflects both range-dependent signal attenuation and aspect-angle-dependent scattering. To enable a direct comparison between the physical model output and the \ac{ai}-derived features, this intensity estimate is normalized per edge. The resulting $I_i^{\mathrm{norm}}$ serves as a position-dependent weighting factor in $[0,1]$ that scales the nominal detection count $N_i$ according to the relative reflectivity of each bin within the corresponding edge zone. Since radar intensity values span several orders of magnitude, a logarithmic transformation is applied before normalization to compress the dynamic range and prevent dominant bins from disproportionately influencing the comparison:
\begin{align}
    I_i^{\mathrm{norm}} =
    \frac{\log(I_i) - \min_{j}(\log(I_j))}
    {\max_{j}(\log(I_j)) - \min_{j}(\log(I_j))},
\end{align}
where $j$ runs over all bins of the corresponding edge zone and 
$I_i^{\mathrm{norm}} \in [0,1]$.

The angular weighting function $w(\theta_i)$ modulates the detection count based on the aspect-dependent detection probability and is composed of two multiplicative terms:
\begin{align}
    w(\theta_i) =
    \underbrace{\bigl(\alpha + (1-\alpha)\sin\theta_i\bigr)}_{\text{sinusoidal term}}
    \cdot
    \underbrace{\left(\beta + (1-\beta)
    \frac{\theta_i - \theta_{\min}}
    {\theta_{\max}-\theta_{\min}}\right)}_{\text{linear term}}.
\end{align}
The sinusoidal term captures the aspect-angle dependence of radar detectability, reflecting the physical observation that the number of detections along a surface increases as the angle of incidence deviates from broadside toward grazing incidence. The linear term introduces a gradual increase in detection probability across the normalized angular range, capturing the growth from the minimum to the maximum observed aspect angle within each edge. Here, $\theta_{\min}$ and $\theta_{\max}$ denote the minimum and maximum aspect angles across all bins of the corresponding edge, normalizing $\theta_i$ to $[0,1]$. The parameters $\alpha$ and $\beta$ control the minimum detection weight at broadside incidence and the lower angular boundary, respectively, and are determined empirically based on the physical characteristics of the radar sensor used in this work, as described in Section~\ref{sec:empirical}.

The binary indicator $f_{\mathrm{FOV}} \in \{0,1\}$ excludes bins that lie outside the radar field of view, ensuring that only bins within the sensor's angular coverage contribute to the detection count. 

Together with $N_i$, $I_i^{\mathrm{norm}}$, and $f_{\mathrm{FOV}}$, all factors are combined into the per-bin detection count $D_i$ as defined in~\eqref{eq:detection_count}.

To account for multipath reflections, which can produce additional detections at positions not directly associated with the vehicle surface, an additive contribution is included:
\begin{align}
    D_i^{\mathrm{final}} = D_i + \alpha_m h_i,
\end{align}
where $h_i$ denotes the expected multipath detection count at bin $i$, which acts as the first reflection point of the multipath signal, and $\alpha_m$ is a distance-dependent attenuation factor accounting for the increased path length of the multipath signal, as longer travel distances result in a weaker received signal.

The resulting $D_i^{\mathrm{final}}$ integrates geometric visibility, scattering intensity, angle-dependent detection probability, and multipath contributions into a unified per-bin detection count that serves as the reference distribution for the subsequent computation of the Wasserstein distance.

\subsection{Rotation Score}
\label{sec:rotation_correction}
The proposed correction method evaluates the reliability of the \ac{ai}-based yaw estimate by comparing radar-derived features against physical model predictions through an aggregated score. Based on this score, the optimal rotation candidate is selected and used to correct the \ac{ai} heading estimate. For each frame $f$, the processing within the rotation score block is described in the following steps.

\subsection*{Step~1: Rotation Candidates}
A discrete set of rotation candidates
\[
  \Theta = \{-10^\circ,\,-9^\circ,\,\ldots,\,+10^\circ\}
\]
with a step size of \SI{1}{\degree} is defined. The rotation search loop iterates over each $\theta \in \Theta$, rotating the \ac{ai} prediction by $\theta$ and forwarding it through the physical model as well as the spatial quantizer, whose output distributions are then evaluated in the rotation score block in Steps~\num{2}--\num{4}. This process repeats for all candidates before the optimal rotation is selected in Step~\num{5}.

\subsection*{Step~2: Distribution Alignment}
For each rotation $\theta$, the feature profiles extracted from the spatial quantizer and the physical model are aligned for comparison. Three complementary feature types are considered: the Euclidean distance (\textit{Euclid}) captures the spatial displacement between the observed radar detections and the predicted edge positions, the detection count (\textit{DetCount}) quantifies the local measurement density per spatial bin, and the reflection intensity (\textit{Intensity}) encodes the radar backscatter strength along each edge. The observed distributions from the spatial quantizer and the corresponding distributions from the physical model are paired along the four vehicle sides $s \in \{\text{left, right, front, rear}\}$ and three edge zones $e \in \{e_1, e_2, e_3\}$ for subsequent Wasserstein-based comparison.

\subsection*{Step~3: Wasserstein-Based Scoring}
The agreement between each pair of observed and reference distributions is quantified using the first-order Wasserstein distance $W_1$ \cite{Wasserstein-Earth}, computed on normalized discrete probability distributions:
\begin{align}
  W_z^{(s,e)}(f,\theta)
  &= W_1\!\left(
    \frac{p_z^{(s,e)}}{\bigl\|p_z^{(s,e)}\bigr\|_1},\;
    \frac{q_z^{(s,e)}}{\bigl\|q_z^{(s,e)}\bigr\|_1}
  \right), \notag\\
  &\quad z \in \{\text{Euclid, DetCount, Intensity}\},
\end{align}
where $p_z^{(s,e)}$ denotes the observed feature distribution from the spatial quantizer and $q_z^{(s,e)}$ the corresponding distribution from the physical model, both for feature type $z$, vehicle side $s$, and edge zone $e$. The division by the respective $\ell_1$-norm ensures that both distributions are normalized to unit mass prior to comparison.
The resulting distances are averaged over all sides $s$ and edge zones $e$ to obtain a single distance value per feature type and rotation candidate $\theta$:
\begin{equation}
  \bar{W}_z(f,\theta)
  = \frac{1}{|\mathcal{S}|\,|\mathcal{E}|}
  \sum_{s\,\in\,\mathcal{S}}\sum_{e\,\in\,\mathcal{E}}
  W_z^{(s,e)}(f,\theta),
\end{equation}
where $|\mathcal{S}|$ and $|\mathcal{E}|$ denote the total number of vehicle sides and edge zones, respectively. To enable a fair comparison across feature types with different scales, each averaged distance is normalized frame-wise across all rotation candidates $\Theta$ to $[0,1]$:
\begin{equation}
  \tilde{W}_z(f,\theta)
  = \frac{
    \bar{W}_z(f,\theta) - \displaystyle\min_{\theta'}\,\bar{W}_z(f,\theta')
  }{
    \displaystyle\max_{\theta'}\,\bar{W}_z(f,\theta')
    - \displaystyle\min_{\theta'}\,\bar{W}_z(f,\theta')
  },
\end{equation}
where the minimum and maximum are taken over all rotation candidates $\theta' \in \Theta$ for the current frame $f$.

\subsection*{Step~4: Score Aggregation}
The three normalized distances are summed with equal weights to obtain a scalar consistency score:
\begin{equation}
  S(f,\theta)
  = \tilde{W}_{\mathrm{Euclid}}(f,\theta)
  + \tilde{W}_{\mathrm{DetCount}}(f,\theta)
  + \tilde{W}_{\mathrm{Intensity}}(f,\theta),
  \label{eq:score}
\end{equation}
where $S(f,\theta) \in [0,3]$ since each normalized term lies in $[0,1]$. A low score indicates strong agreement between the rotated \ac{ai} prediction and the physical model, while a high score indicates a poor match. Equal weighting was applied to all three features so that none would be given priority over the others.

\subsection*{Step~5: Optimal Rotation}
After all candidates in $\Theta$ have been evaluated, the rotation yielding the minimum aggregated score is selected as the optimal yaw correction for frame $f$:
\begin{equation}
  \theta^*(f)
  = \arg\min_{\theta\,\in\,\Theta}\; S(f,\theta),
  \label{eq:argmin}
\end{equation}
where $\theta^*(f) \in \Theta$ denotes the candidate rotation that best aligns the bounding box with the observed radar point cloud. The correction $\theta^*(f)$ is added to the original \ac{ai} yaw prediction $\psi^{\text{AI}}$, yielding the final corrected vehicle orientation
\begin{equation}
  \psi^*(f) = \psi^{\text{AI}} + \theta^*(f).
\end{equation}

\section{Measurements}
\label{sec:measure}

\subsection{Radar}
\label{sec:radar}
The measurements were acquired using an INRAS Radarlog sensor operating at \SI{76}{\giga\hertz} with \ac{fmcw} chirp sequence modulation. The system was configured with a bandwidth of \SI{1.5}{\giga\hertz} (\SIrange{76}{77.5}{\giga\hertz}), a chirp duration of \SI{51.2}{\micro\second}, and a chirp repetition rate of \SI{60}{\micro\second}, encompassing \num{512} chirps per coherent processing interval. This configuration yields a range gate length of \SI{0.1}{\meter}, a maximum unambiguous range of \SI{25}{\meter}, and a Doppler resolution of \SI{0.06}{\meter\per\second}. The radar front-end comprises a \ac{ula} with one transmit and \num{16} receive antennas, providing an azimuth angular accuracy of \SI{0.12}{\degree}. During data collection, the sensor remained stationary at a height of \SI{0.3}{\meter}, positioned at the center of the testing facility.

\subsection{Empirical Parameter}
\label{sec:empirical}
The physical model requires two \ac{rcs}-related parameters, $\Sigma_{\text{wheel}}$ and $\Sigma_{\text{chassis}}$, which represent the mean \ac{rcs} of the wheel and chassis regions, respectively. These parameters are determined empirically from the validation dataset to ensure they reflect the actual reflectivity characteristics.
For this purpose, the radar point cloud of the validation dataset is decomposed into two disjoint subsets: a wheel point cloud $\mathrm{pc}_{\text{wheel}}$ and a chassis point cloud $\mathrm{pc}_{\text{chassis}}$. The extraction procedure is described in detail in~\cite{zaumseil_YOLO}.

The parameters $\alpha$ and $\beta$ were determined by a grid search over the validation set, minimizing the deviation between the model-predicted intensity distribution and the measured radar intensity profile along each edge. The resulting values $\alpha = 0.7$ and $\beta = 0.85$ are specific to the radar sensor in this work.

Substituting the sensor-specific parameters from Section~\ref{sec:radar} and the bin area $A_{\text{bin}} = \SI{10}{\centi\metre} \times \SI{10}{\centi\metre} = \SI{0.01}{\metre\squared}$ into \eqref{eq:N0}, the nominal detection count evaluates to
\begin{align}
    N_0(R_i) \approx \frac{4.78}{R_i}.
\end{align}

\subsection{Datasets}
\label{sec:data}

\begin{figure}[t]
    \centering
    \subfloat[Validation dataset]
    {\includegraphics[width=0.95\linewidth]{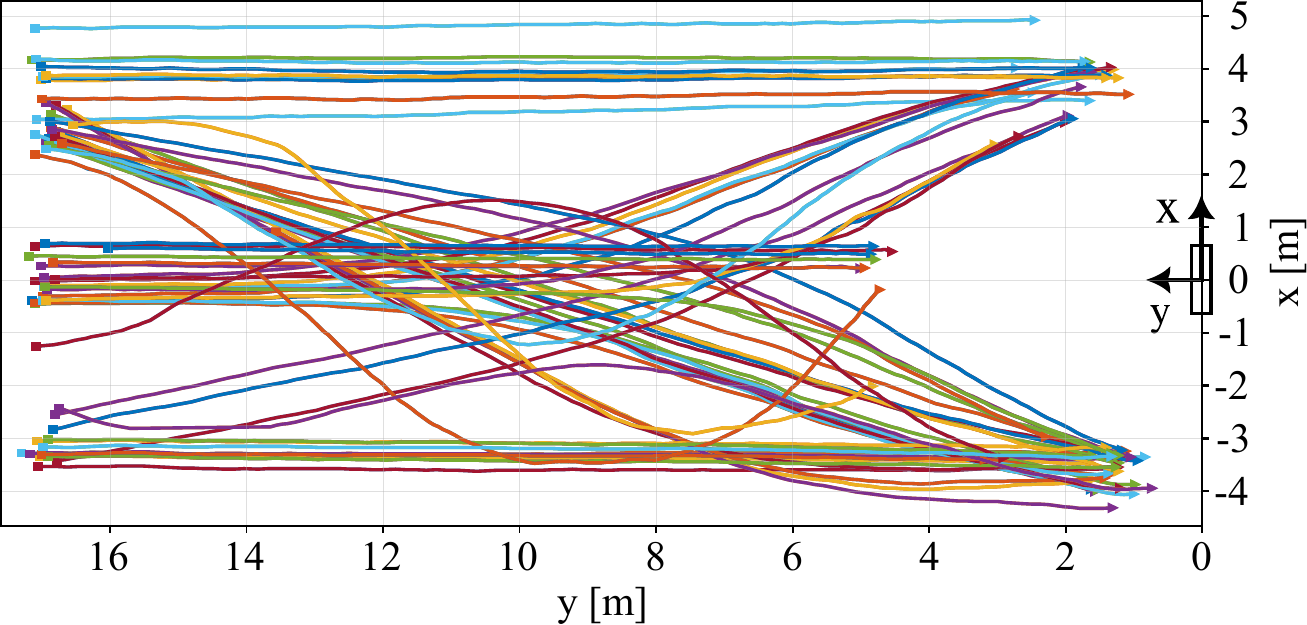}
    \label{fig:valdi_dataset}}
    \\
    \subfloat[Test dataset]
    {\includegraphics[width=0.95\linewidth]{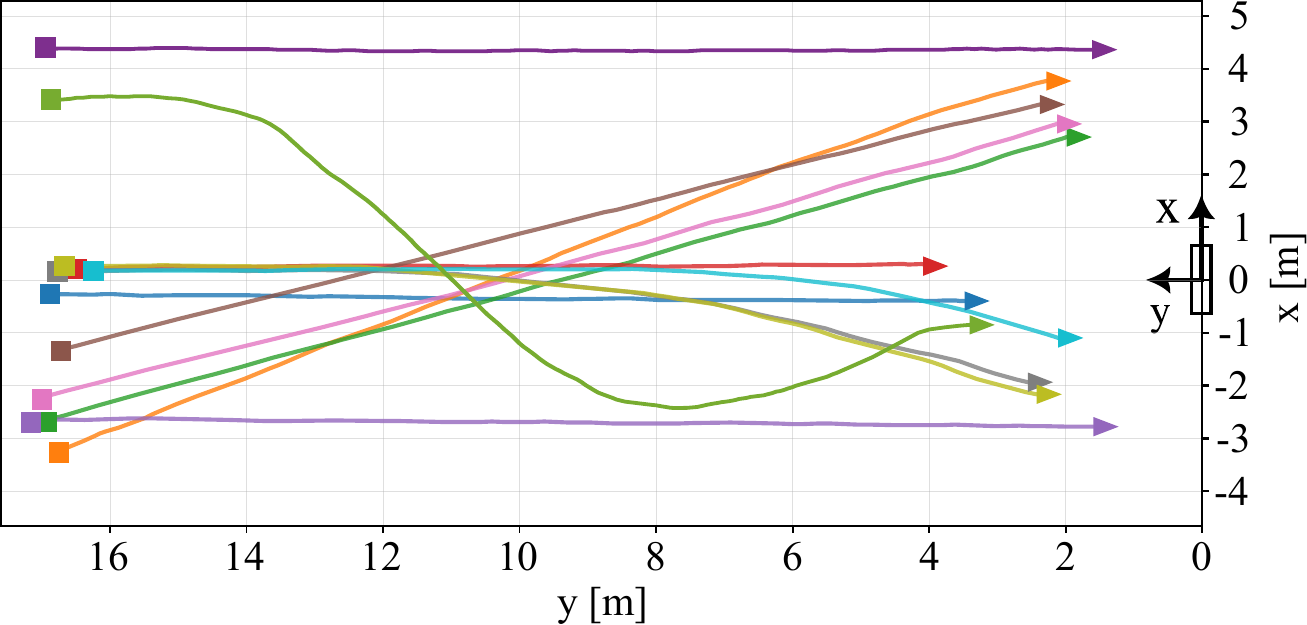}
    \label{fig:test_dataset}}
    \caption{Measurement trajectories used for model validation and evaluation. Start (\textcolor{black}{\rule{0.5em}{0.5em}}) and end (\textbf{$\blacktriangledown$}) positions are marked accordingly.}
    \label{fig:datasets}
\end{figure}

The measurement database used in this work is described in detail in~\cite{zaumseil_YOLO} and summarized here for completeness. A dedicated high-resolution radar database was recorded under controlled indoor conditions, focusing on a single test vehicle to ensure consistent \ac{rcs} characteristics and repeatable \ac{md} signatures~\cite{zaumseil_YOLO}. The measurements were conducted using predefined trajectories with linear and dynamic vehicle passes, including variations in overlap, braking maneuvers, and initial approach angles, as illustrated in Figs.~\ref{fig:valdi_dataset} and~\ref{fig:test_dataset}. In total, \num{92} measurement runs were conducted, each comprising approximately \num{120} frames.

The complete database is divided into a validation dataset and a test dataset. The validation dataset is identical to the dataset used to train the ComplexYOLO model in~\ref{sec:ki-model}, comprising \num{7324} frames, and is used here exclusively to validate the physical radar model and to determine the empirical parameters described in Section~\ref{sec:empirical}. The remaining measurements constitute the test dataset of \num{1192} frames, which is used exclusively to evaluate the proposed physical model and was not used during its development.
\section{Results}
\label{sec:results}

\begin{figure}[b]
    \centering
    \includegraphics[width=0.95\linewidth]{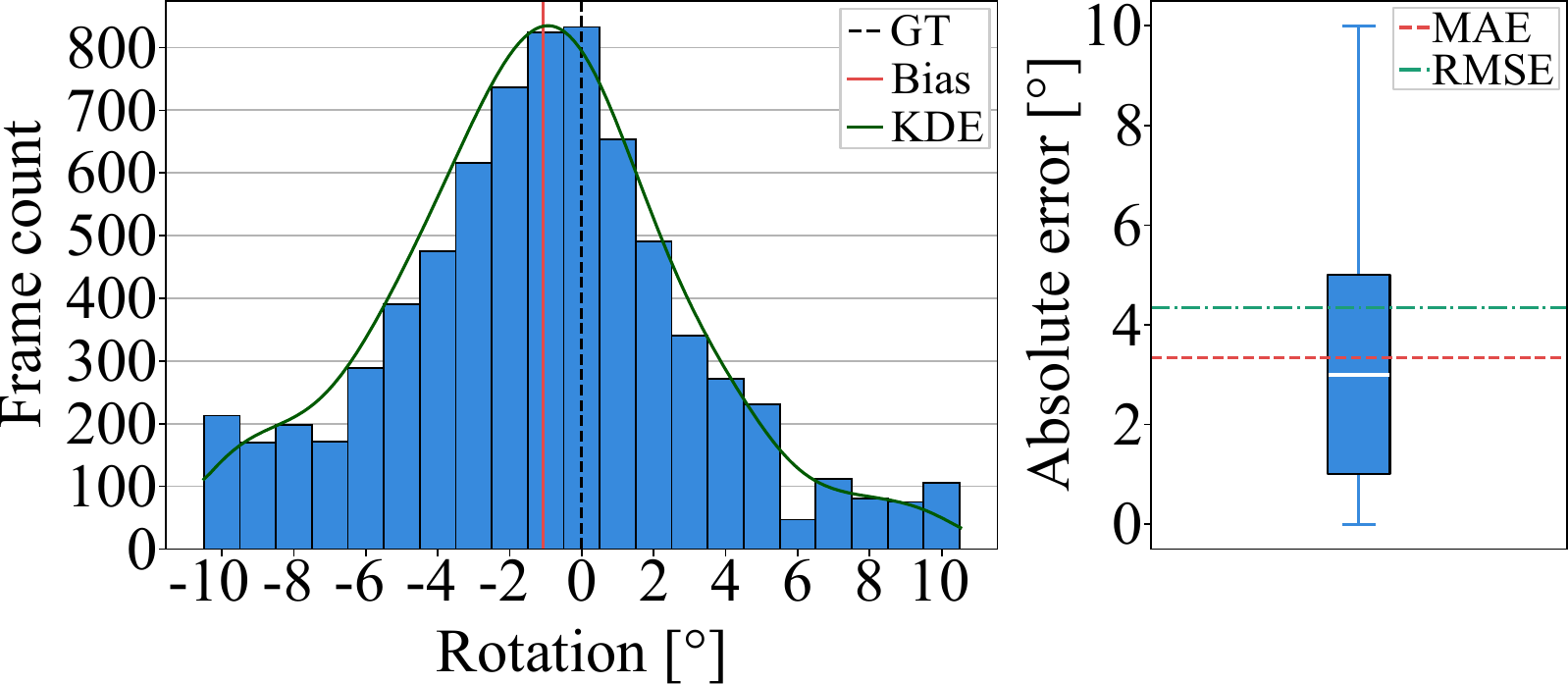}
    \caption{Evaluation of the physical model on the validation set, in which the model is applied directly to the ground-truth object state to assess its intrinsic correction capability. The left plot shows the distribution of rotation corrections overlaid with an \acf{kde}, with the ground-truth reference at \SI{0}{\degree} (black dashed line). The right plot shows the absolute error distribution as a boxplot with \ac{mae} and \ac{rmse} indicated as reference lines.}
    \label{fig:rotation_validation}
\end{figure}

The physical model is first evaluated on the validation set of \num{7324} frames, where it is applied directly to the ground truth object state to assess its intrinsic correction capability. The ground truth was recorded using a reference measurement system, providing a reliable basis for quantitative evaluation. As shown in Fig.~\ref{fig:rotation_validation}, the distribution of predicted rotation corrections is approximately symmetric around the ground truth at \SI{0}{\degree}, with a slight negative bias of \SI{-1.07}{\degree}, a \ac{mae} of \SI{3.35}{\degree}, and an \ac{rmse} of \SI{4.34}{\degree}.

\begin{figure}[t!]
    \centering
    \includegraphics[width=1\linewidth]{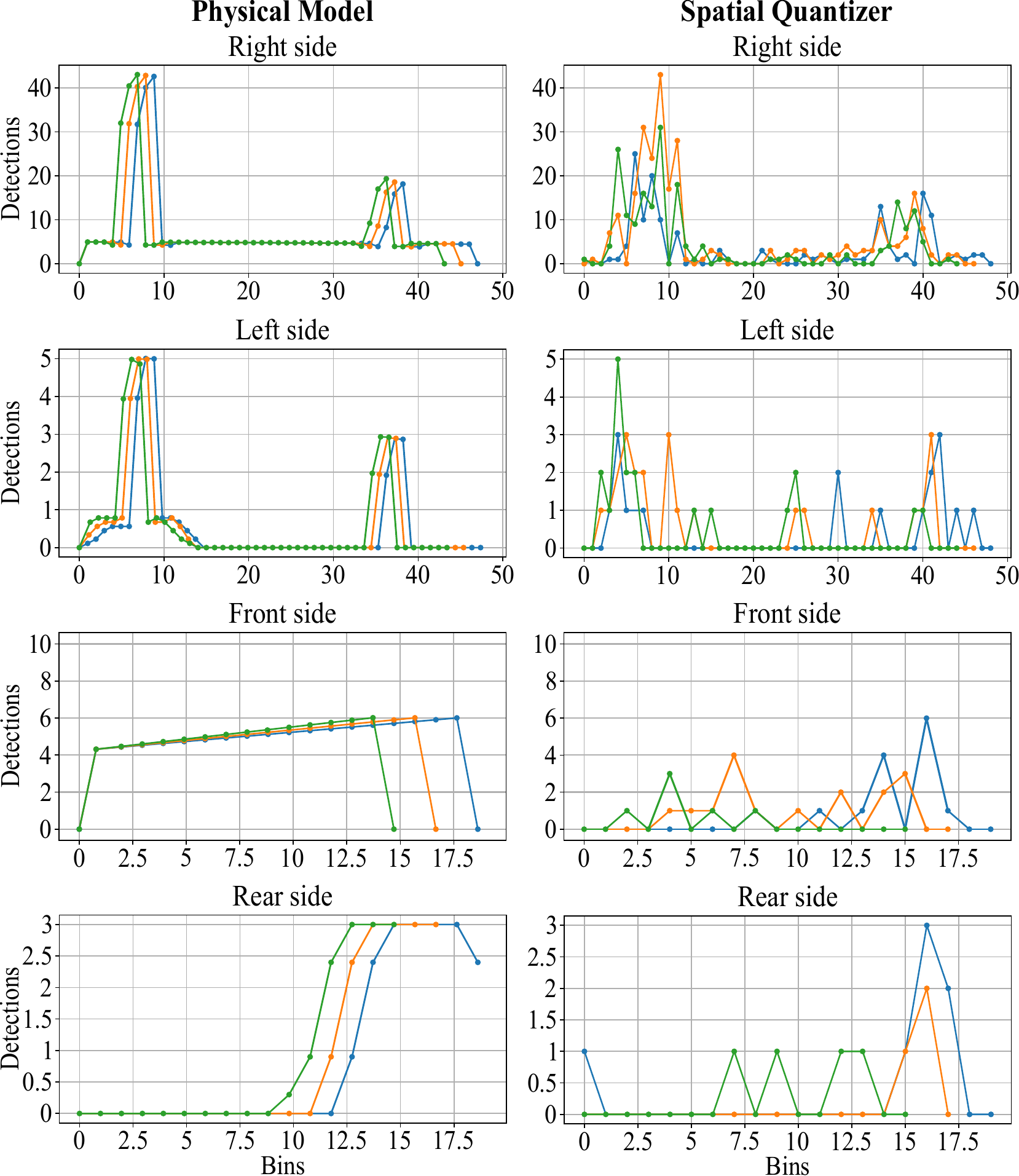}
    \caption{The left column shows the \textit{Physical Model} output, and the right column the corresponding distributions derived from the rotated bounding box projected onto the measured radar point cloud via the \textit{Spatial Quantizer}. Edge zones $e_1$, $e_2$, and $e_3$ are shown in blue, orange, and green, respectively.}
    \label{fig:detection_comparison}
\end{figure}

Fig.~\ref{fig:detection_comparison} provides a direct comparison between the two representations that form the basis of the rotation scoring: the distributions generated by the physical model and the corresponding distributions obtained by projecting the rotated bounding box onto the measured radar point cloud via the spatial quantizer, shown for all four bounding box sides and three edge zones per side. The physical model captures the dominant spatial structure of the radar detection pattern across all four sides and correctly locates the principal detection peaks identified by the spatial quantizer, such as the left and right side maxima. This agreement reflects that the detection count $N_i$ directly captures the range-dependent detection density that drives the observed peaks in the measured point cloud. The physical model produces a smoother profile than the spatial quantizer because it is based on a closed-form formulation rather than discrete bin counts, which are subject to bin-to-bin noise inherent in spatial discretization. The close alignment between the three edge zones within the physical model further indicates that this trend is robust and not an artifact of a single candidate selection.

\begin{figure}[t]
    \centering
    \includegraphics[width=0.95\linewidth]{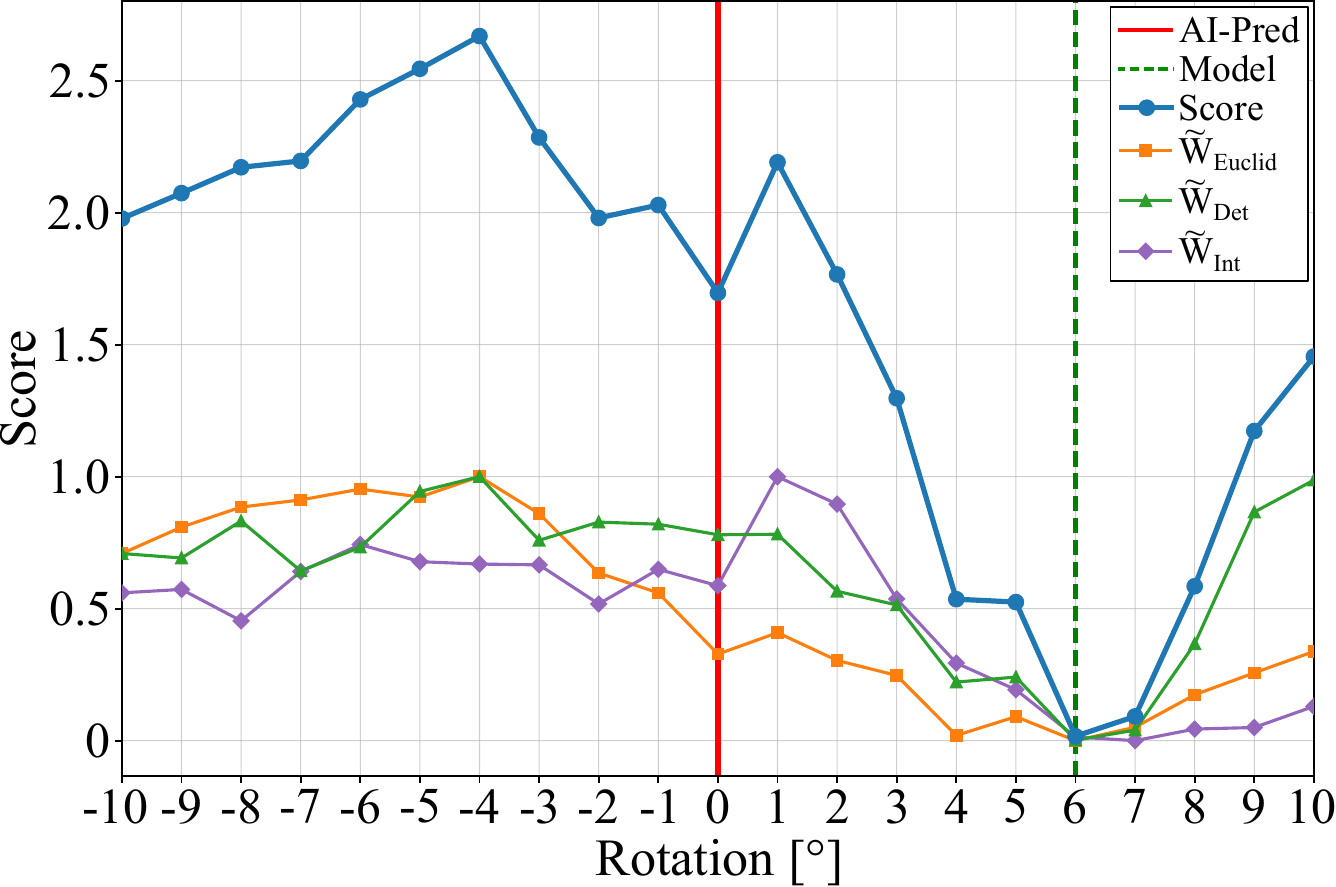}
    \caption{Rotation scoring curves for a single exemplary test frame. The normalized Wasserstein distances $\tilde{W}_\text{Euclid}$, $\tilde{W}_\text{Det}$, and $\tilde{W}_\text{Int}$ and the aggregated score $S$ are shown over candidate rotations from \SI{-10}{\degree} to \SI{10}{\degree}. The green dashed line (\textit{Physical Model}) marks the selected optimal rotation, and the red vertical line (\textit{AI-Pred}) marks the \ac{ai} prediction.}
    \label{fig:rotation_scoring}
\end{figure}

\begin{figure}[b]
    \centering
    \includegraphics[width=0.55\linewidth]{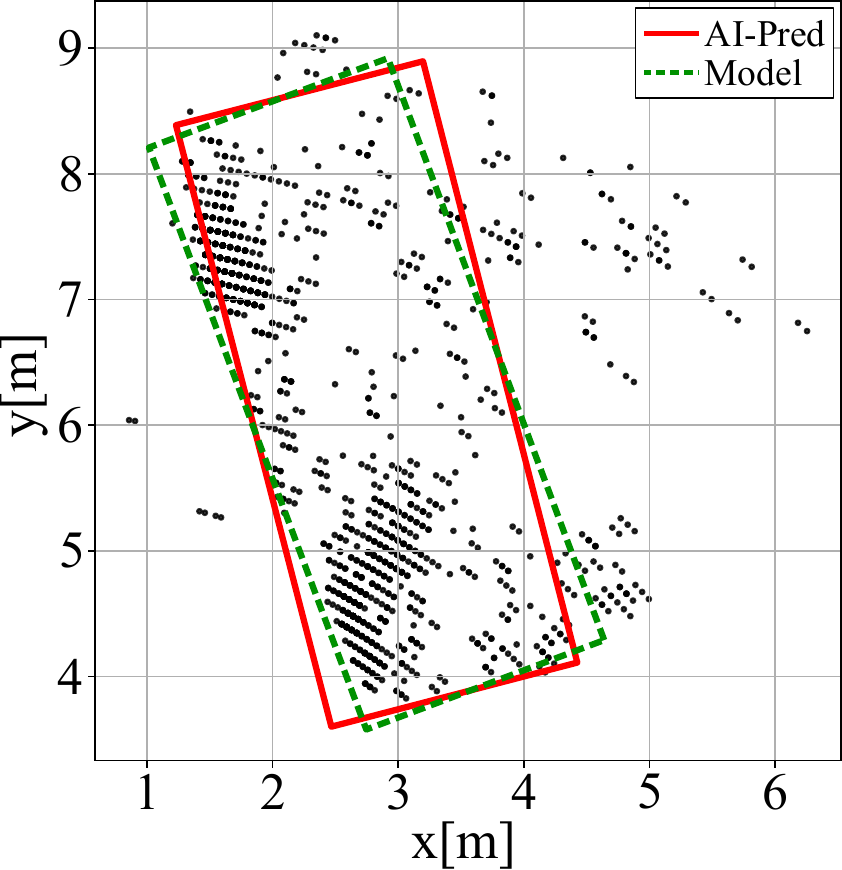}
    \caption{Bounding box alignment for the frame corresponding to Fig.~\ref{fig:rotation_scoring}. The red solid box represents the original \ac{ai} heading prediction, and the green dashed box the corrected orientation after applying the model rotation.}
    \label{fig:bounding_box_alignment}
\end{figure}

To further illustrate the physics-based assessment framework's behavior on a single test frame, Fig.~\ref{fig:rotation_scoring} shows the normalized Wasserstein distances $\tilde{W}_\text{Euclid}$, $\tilde{W}_\text{Det}$, and $\tilde{W}_\text{Int}$ together with the aggregated score $S$ over the full candidate rotation range. All three components consistently reach their minimum around \SI{6}{\degree}, where $S$ also attains its global minimum, thereby identifying this as the optimal correction. Notably, the \ac{ai}-predicted heading at \SI{0}{\degree} corresponds to a comparatively high score of $S \approx 1.7$, well above the minimum, indicating a substantial discrepancy between the \ac{ai} prediction and the physically optimal yaw for this frame. The score exhibits a local minimum near \SI{1}{\degree} before rising again toward \SI{-4}{\degree}, suggesting that the underlying detection geometry is not strictly unimodal and that a coarse search could converge to a suboptimal local minimum if not evaluated over the full candidate range. The three component distances are not perfectly aligned in their individual minima, particularly $\tilde{W}_\text{Int}$, which peaks sharply around \SI{1}{\degree}, but their combination in $S$ resolves this ambiguity and consistently points to \SI{6}{\degree} as the global optimum.

The alignment of the corresponding bounding box is shown in Fig.~\ref{fig:bounding_box_alignment}: the corrected orientation at \SI{6}{\degree} (green dashed box) visibly improves the spatial fit compared to the original \ac{ai} prediction at \SI{0}{\degree} (solid red box). The corrected box more closely follows the elongated point cloud structure of the detected vehicle, while the original prediction is visibly rotated relative to the main axis of the detection cluster. The angular offset between both boxes increases toward the far end of the bounding box, consistent with a rotational misalignment rather than a translational one. The detections visible outside and behind the bounding box are due to multipath reflections.

\begin{figure}[t]
    \centering
    \includegraphics[width=0.95\linewidth]{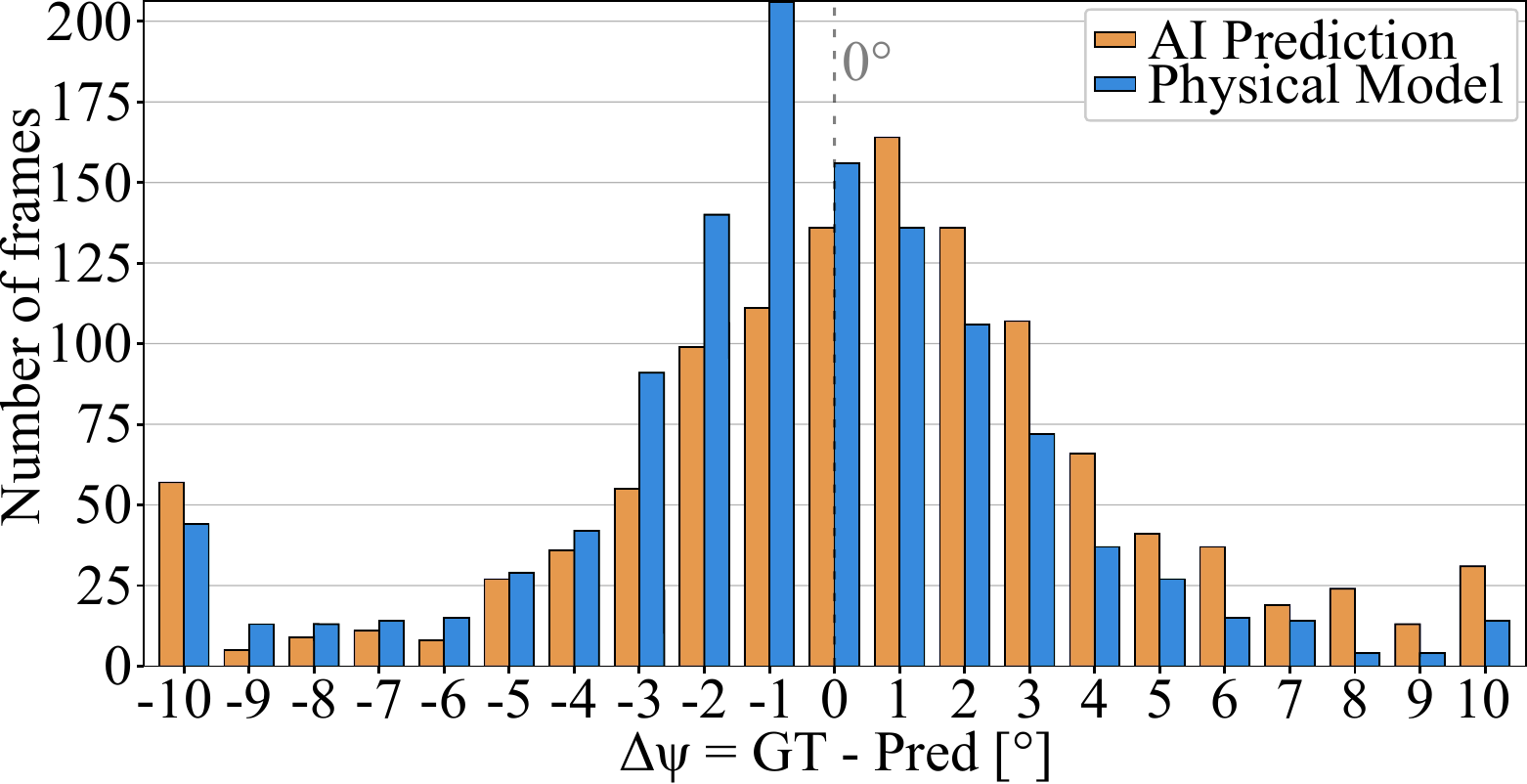}
    \caption{Comparison of the yaw angle error distributions $\Delta\psi = \psi_{\text{GT}} - psi_{\text{Pred}}$ for the \ac{ai}-only baseline (\textit{AI Prediction}) and the proposed physics-based assessment framework (\textit{Physical Model}) over \num{1192} test frames.}
    \label{fig:results_histo_test}
\end{figure}

The physics-based assessment framework is subsequently applied to correct the \ac{ai} predictions on the held-out test set of \num{1192} frames. Fig.~\ref{fig:results_histo_test} compares the yaw angle error distributions $\Delta\psi = \psi_\text{GT} - \psi_\text{Pred}$ for the \ac{ai}-only baseline and the proposed fusion of \ac{ai} prediction and physical model correction. The combined approach clearly shifts the error mass toward zero. The standard deviation is reduced by \SI{19}{\percent} from $\sigma_\text{AI} = \SI{5.27}{\degree}$ to $\sigma_\text{Model} = \SI{4.24}{\degree}$, while the systematic bias remains close to zero.

\begin{table}[b]
    \centering
    \caption{Comparison of heading prediction errors between the original \ac{ai} model and the proposed correction model across the test frames.}
    \label{tab:heading_error_comparison}
    \begin{tabular}{lrrrr}
    \toprule
    \textbf{Metric} & \textbf{Unit} & \textbf{AI-Pred} & 
    \textbf{Model} & \textbf{$\Delta$} \\
    \midrule
    RMSE        & $^\circ$ & \num{5.28}  & \num{4.29}  & \num{-0.99} \\
    Mean AE     & $^\circ$ & \num{3.50}  & \num{2.84}  & \num{-0.66} \\
    Median AE   & $^\circ$ & \num{2.24}  & \num{1.84}  & \num{-0.40} \\
    \midrule
    P90         & $^\circ$ & \num{7.93}  & \num{6.50}  & \num{-1.43} \\
    P95         & $^\circ$ & \num{11.16} & \num{9.37}  & \num{-1.79} \\
    \midrule
    Frames $\phantom{\SI{2}{\degree} \leq{}} |\Delta\psi| < \SI{2}{\degree}$     & \si{\percent} & \num{34.5} & \num{41.8} & \num{+7.3} \\
    Frames $\SI{2}{\degree} \leq |\Delta\psi| < \SI{5}{\degree}$                 & \si{\percent} & \num{41.8} & \num{40.9} & \num{-0.9} \\
    Frames $\phantom{\SI{2}{\degree} \leq{}} |\Delta\psi| \geq \SI{5}{\degree}$  & \si{\percent} & \num{23.7} & \num{15.9} & \num{-7.7} \\
    \midrule
    Jitter      & $^\circ$ & \num{5.24}  & \num{4.28}  & \num{-0.95} \\
    Bias        & $^\circ$ & \num{-0.36} & \num{0.68}  & \num{0.32} \\
    Std. Dev.   & $^\circ$ & \num{5.27}  & \num{4.24}  & \num{-1.04} \\
    \bottomrule
    \end{tabular}
\end{table}

The quantitative metrics are summarized in Tab.~\ref{tab:heading_error_comparison}. The most notable improvements are observed at both ends of the error spectrum. In the high-precision regime, the proportion of frames with $|\Delta\psi| < \SI{2}{\degree}$ increases from \SI{34.5}{\percent} to \SI{41.8}{\percent}, an improvement of 7.3 percentage points. In the large-error regime, the proportion of frames with $|\Delta\psi| \geq \SI{5}{\degree}$ decreases from \SI{23.7}{\percent} to \SI{15.9}{\percent}, a reduction of 7.7 percentage points, indicating improved robustness in challenging frames. The \ac{rmse} is reduced from \SI{5.28}{\degree} to \SI{4.29}{\degree}, and the \num{95}th percentile drops from \SI{11.16}{\degree} to \SI{9.37}{\degree}. The improvement across the full error distribution is statistically significant according to the Wilcoxon signed-rank test, one-sided, $W = \num{313410}$, $p = \num{9.48e-19} \ll 0.001$.

\section{Discussion}
\label{sec:discussion}

The results presented in Section~\ref{sec:results} demonstrate that the proposed radar-based physical model consistently improves the estimation of the \ac{ai} prediction across all evaluated metrics. In the following, the key findings are interpreted in the broader context of radar-based vehicle perception, and the limitations of the current approach are discussed.

\subsection{Interpretation of Results}
The consistent reduction in \ac{mae}, \ac{rmse}, and standard deviation of approximately \SI{19}{\percent} confirms that the physical radar model provides complementary information that \ac{ai} predictions alone do not fully capture. Particularly noteworthy is the improvement in the tails of the error distribution: the proportion of frames with large errors $|\Delta\psi| \geq \SI{5}{\degree}$ is reduced by \SI{7.7}{\percent} from \SI{23.7}{\percent} to \SI{15.9}{\percent}, indicating that the correction model effectively suppresses outlier predictions that could be critical in safety-relevant applications such as \ac{aeb} or airbag deployment. At the same time, the proportion of highly accurate predictions with $|\Delta\psi| < \SI{2}{\degree}$ increases by \SI{7.3}{\percent} from \SI{34.5}{\percent} to \SI{41.8}{\percent}, reflecting an overall shift of the error mass toward zero.

The \ac{ai} model provides a strong initial estimate, while the physical scoring based on Wasserstein distances refines it by aligning the predicted bounding box with the underlying radar point cloud distribution. This demonstrates that physics-based reasoning can provide complementary information that data-driven models alone do not fully capture, particularly when the \ac{ai} prediction deviates from the physically expected radar response. The statistical significance of the improvement, confirmed by the Wilcoxon signed-rank test ($p = \num{9.48e-19}$), further validates that the observed gains are not attributable to random variation but reflect a systematic improvement introduced by the correction model.

\subsection{Robustness and Generalization}
The evaluation on the validation set assesses the model's intrinsic correction capability, while the held-out test set provides a more rigorous assessment of performance across unseen scenarios. The consistent improvement across all test trajectories, covering the principal driving patterns shown in Fig.~\ref{fig:test_dataset}, confirms that the physics-based formulation is applicable across varying object configurations without scenario-specific adaptation. 
Nevertheless, the quality of the correction depends on the informative content of the radar response for a given object configuration. In scenarios where only a single edge is visible, such as a pure frontal view without wheel exposure, the available feature distributions may be less distinctive, reducing the achievable correction accuracy. Similarly, in lateral driving configurations, the Doppler contrast between individual edges decreases, which may reduce the discriminative value of velocity features. These configuration-dependent variations in feature informativeness represent the primary source of residual uncertainty and warrant further investigation.

\subsection{Bias and Systematic Tendencies}
A slight negative bias of \SI{-1.07}{\degree} is observed in the validation set evaluation, and the bias shifts from \SI{-0.36}{\degree} to \SI{+0.68}{\degree} on the test set after correction. While both values remain small in absolute magnitude, the sign change suggests that the correction model introduces a minor systematic tendency in the opposite direction. This may be attributed to a systematic asymmetry in the training data distribution and warrants consideration in future refinements.

\subsection{Limitations and Future Work}
The current parameterization relies on empirically determined values that were tuned to the specific sensor used in this work. A systematic calibration procedure or a learned parameter optimization could improve transferability to other radar configurations.
\section{Conclusion}
\label{sec:conclusion}

This work presents a physics-based plausibility module to validate and correct \ac{ai}-predicted vehicle headings in automotive radar perception. The proposed cascaded pipeline combines a data-driven bounding box estimate with a Wasserstein-distance-based rotation score derived from a physical radar model, yielding statistically significant improvements across all evaluated metrics on \num{1192} test frames. The results demonstrate that physics-based reasoning can meaningfully complement \ac{ai} predictions in safety-critical perception tasks without requiring additional training data or model retraining. Future work will investigate temporal consistency constraints, systematic parameter calibration, and improved handling of lateral driving scenarios to further enhance robustness and transferability to other radar configurations.

\bibliographystyle{IEEEtran}
\bibliography{Bibliography/Reference}

\end{document}